%% file: main.tex
\documentclass[conference]{IEEEtran}
\IEEEoverridecommandlockouts
\usepackage[utf8]{inputenc}
\usepackage[T1]{fontenc}
\usepackage{amsmath,amssymb,amsthm}
\usepackage{graphicx}
\usepackage{booktabs}
\usepackage{xcolor}
\usepackage{tikz}
\usetikzlibrary{arrows.meta,positioning}
\usepackage{cite}
\usepackage[colorlinks=true,linkcolor=blue!60!black,citecolor=blue!60!black,urlcolor=blue!60!black]{hyperref}
\usepackage{adjustbox}

\newtheorem{theorem}{Theorem}
\newtheorem{proposition}{Proposition}
\newtheorem{lemma}{Lemma}

\begin{document}

\title{Decentralised Drone Swarm Formation under Silent Failures with ACO Pheromone Claims and PSO Tuning}

\author{\IEEEauthorblockN{Md.\ Abu Horaira}
\IEEEauthorblockA{Department of Computer Science and Engineering\\University of Dhaka, Bangladesh\\mdabu-2021911202@cs.du.ac.bd}}

\maketitle

\input{abstract}


\input{sec_intro}
\input{sec_method}
\input{sec_setup}
\input{sec_results}
\input{sec_discussion}

\section*{Code and Data Availability}
\label{sec:artefacts}
The simulator, the browser demonstration, all experiment scripts, the raw per-run results and the scripts that produce every table and figure are released as ancillary files of this arXiv submission (\texttt{anc/dronesim/}). A single command, \texttt{sh exp/run\_all.sh}, reproduces all results on a desktop CPU: about five hours on eight cores, plus about one hour for each tuning replication. In the demonstration, which runs the same simulator code in a browser, the user can destroy or add drones at any time, toggle the safety filter and inspect the claim table of any drone.

\section*{Acknowledgment}
AI-assisted tools were used to develop the simulation code and the experiment and analysis scripts, and to draft the manuscript. The author reviewed and verified all content, code, results and references, and takes full responsibility for the paper.

\bibliographystyle{IEEEtran}
\bibliography{references}

\clearpage
\input{sec_appendix}

\end{document}

%% file: abstract.tex
\begin{abstract}
We study decentralised drone formation, without a central planner, under the conditions that make it hard: nobody knows how many drones are alive, failures are silent, radio is short-ranged and lossy, and actuation is noisy. In our controller, ant colony optimisation (ACO) allocates drones to picture positions through claims that behave as reinforced, evaporating pheromone and are gossiped between neighbours, so evaporation doubles as failure detection. A particle swarm optimisation (PSO) law flies the drones, a noise-buffered Voronoi filter keeps them apart, and PSO also tunes the controller offline. We prove that claims of destroyed drones vanish within bounded time, that conflicts between drones that exchange tables are resolved consistently, that survivors cover the picture to within a factor of two, and that drones cannot collide while their safety cells are feasible. On held-out seeds and with two tuned configurations, the controller formed the picture in 95--99\% of runs in the hardest regime, restored it after a silent loss of half the swarm in 96--100\% of runs at normal difficulty, transferred unchanged to eight unseen shapes and to 300 drones, and overlapped in 5 of 21\,088 runs. A CBBA-style baseline that shares the evaporating claims but uses argmax without swaps was at least as reliable as ours once tuned with the same budget, and repaired faster, whereas ours generalised better to unseen shapes. The benefit thus comes from the pheromone dynamics and from tuning, not from ACO's stochastic construction or PSO's social terms. A centralised oracle is faster.
\end{abstract}

%% file: sec_intro.tex
\section{Introduction}

Drone light shows, search-and-rescue teams and sensor deployments need many inexpensive aircraft to take up a prescribed arrangement while individual aircraft fail, join or lose contact. Formation methods that come with guarantees typically plan every assignment and trajectory centrally, from the positions of all drones~\cite{turpin2014capt}. The planner is then a single point of failure and a bandwidth bottleneck. A \emph{decentralised} swarm, in which every drone decides from local information, removes both. It must, however, answer three questions without a coordinator: \emph{who goes where}, \emph{how to fly there}, and \emph{how to keep drones from colliding} on the way.

No single question here is hard on its own. The difficulty is that five conditions hold together in the field:
\begin{itemize}
\item nobody knows how many drones are alive, so any plan that counts the swarm first is outdated after the next loss;
\item failures are silent, because a drone that crashes sends no farewell;
\item radio is short-ranged and lossy, so two drones can believe they own the same position and must settle this without an arbiter;
\item actuation noise moves drones in ways nobody commanded, which eats into every safety margin;
\item every swarm algorithm has parameters, and values that work for one picture or radio condition may fail for another.
\end{itemize}

Swarm intelligence offers a mechanism for most of these conditions. In ant colony optimisation (ACO)~\cite{dorigo1996antsystem,dorigo1997acs,dorigo2004aco}, ants coordinate through pheromone that is reinforced while it is useful and evaporates when it is not. Evaporating pheromone is a memory that fades unless it is refreshed, which is exactly what is needed to notice that a silent drone is gone. Reinforcement lets a well-founded claim prevail over a weaker one without a referee. In particle swarm optimisation (PSO)~\cite{kennedy1995pso,shi1998inertia,clerc2002constriction}, particles move under inertia and are drawn towards good positions found by themselves and their neighbours. This suggests a motion law that smooths noisy steps, and, used offline, a black-box optimiser for the controller's own parameters. Collision avoidance is the one requirement that we deliberately do \emph{not} entrust to swarm intelligence. It must hold even while allocation is wrong, so it sits in an independent geometric filter that comes with a proof.

ACO and PSO are well established for \emph{planning} problems. Their use as the online control law of a physical, failure-prone swarm is much less settled, and the evidence offered for such uses is often thin: one scenario, a few seeds, hand-chosen parameters and weak baselines. We therefore pose four research questions and answer each with controlled experiments on held-out seeds.
\begin{description}
\item[RQ1] Can strictly local, pheromone-based allocation form a prescribed picture as reliably and quickly as established decentralised allocators, and how far is it from a centralised optimum?
\item[RQ2] Does the swarm repair the picture after silent losses, joins and continuous churn, and what limits the speed of repair?
\item[RQ3] Can collisions be excluded while the population changes, and at what cost?
\item[RQ4] Which ingredients of ACO and PSO actually contribute, and does PSO-based tuning generalise beyond its training conditions?
\end{description}

Our contributions are as follows.
\begin{enumerate}
\item \textbf{A controller with three local layers.} The controller combines an ACO allocation layer, in which claims act as reinforced, evaporating pheromone gossiped between neighbours, with a PSO motion layer and a noise-buffered Voronoi safety filter. A priority ordering of the picture lets any number of drones form a recognisable image without counting themselves.
\item \textbf{Analysis.} We prove that ghost claims of destroyed drones vanish from every table within bounded time, whatever the topology or packet loss; that a pairwise conflict is resolved consistently whenever the two drones exchange tables; that every prefix of the priority order covers the sampled picture to within a factor of two of the best possible; that the motion law is stable; and that drones cannot collide under arbitrary losses and joins as long as their safety cells are feasible, a condition that can fail in dense jams. We state where each guarantee ends.
\item \textbf{Evidence.} We compare against a centralised optimal oracle and four decentralised allocators that share our motion and safety layers. The strongest of them, a CBBA-style allocator with reset, is also tuned five times with the same budget as our controller. The experiments cover:
\begin{itemize}
\item twelve shapes, eight of which were never used in design or tuning;
\item four difficulty levels;
\item losses, joins, churn and spare drones;
\item swarms of up to 300 drones;
\item ablations of every component;
\item PSO tuning against random search, five runs per tuned controller under a train/validation/test split;
\item two disjoint sets of test seeds, and two tuned configurations of each tuned controller in the main comparisons between them.
\end{itemize}
The main finding is a negative one about the textbook ingredients: once the baseline is tuned as well, ACO's stochastic construction gives no advantage over deterministic construction, and PSO's social terms none over a damped controller. Run with identical parameters, argmax construction without swaps was at least as reliable in the hardest regime; in benign conditions, whichever rule the parameters had been tuned for was faster. What matters are the evaporating pheromone claims and the offline tuning.
\item \textbf{Artefacts.} The simulator, all experiment code, the raw results and an interactive browser demonstration are released with the paper.
\end{enumerate}

\section{Related Work}

\textbf{Swarm intelligence in robotics.} Bonabeau et al.~\cite{bonabeau1999swarm} and the survey of Brambilla et al.~\cite{brambilla2013swarm} describe stigmergy and self-organisation as the basis of robust, scalable swarms. Rubenstein et al.~\cite{rubenstein2014kilobot} showed that a thousand simple robots can assemble prescribed shapes from local rules and gradient information, with a fixed team of identical robots. Ant robots that lay evaporating, repelling traces cover terrain robustly even when robots fail~\cite{wagner1999evaporating,svennebring2004antrobots}; our claim pheromone uses evaporation in the same coordinating role, but for allocation rather than coverage. We address aerial robots that must also avoid each other and that may be removed or added during the mission.

\textbf{Task allocation.} Gerkey and Matari\'c~\cite{gerkey2004taxonomy} classify multi-robot task allocation. Our setting is single-task robots and single-robot tasks with instantaneous assignment. Dias et al.~\cite{dias2006market} survey market-based methods. The consensus-based bundle algorithm (CBBA)~\cite{choi2009cbba} resolves conflicts by max-consensus on bids and is the standard decentralised baseline; we reproduce its single-assignment core with and without its reset mechanism. Turpin et al.~\cite{turpin2014capt} solve assignment and trajectory planning for formations centrally and with guarantees, and our oracle plays the same role. Pujol-Gonzalez et al.~\cite{pujol2018lorp} and Albani et al.~\cite{albani2021hierarchical} study allocation under limited communication. Pheromone- and market-inspired UAV coordination has recently appeared in search and rescue~\cite{zhu2026mountain,giacomossi2026market}. These works target coverage or task streams rather than a prescribed spatial picture, and none combines claim-pheromone evaporation with a separation filter that comes with a proof. We do not claim primacy; we could not find a head-to-head comparison and therefore built one.

\textbf{Formation control.} Planar shapes have been assigned with Fourier descriptors~\cite{petitprez2023planar}. Flocking drones with optimised local rules have flown outdoors~\cite{vasarhelyi2018flocking}, and learning-based controllers handle dynamic topology~\cite{chen2026madrl}. Reynolds' boids~\cite{reynolds1987boids} introduced separation, alignment and cohesion. These approaches either form shapes with a fixed team, flock without a prescribed picture, or learn a policy that must be trained. Our controller handles a changing team without training, but it does need a template.

\textbf{Collision avoidance.} Reactive separation forces, as in boids, give no guarantee. Reciprocal velocity obstacles (ORCA)~\cite{vandenberg2011orca} and buffered Voronoi cells (BVC)~\cite{zhou2017bvc} give decentralised guarantees under their assumptions. We adopt BVC because its constraint set is a simple intersection of half-planes whose pairwise disjointness yields a short proof, and we extend it with an explicit noise buffer.

\textbf{Parameter tuning.} Hand-set metaheuristic parameters are a known source of unreported degrees of freedom. Eiben and Smit~\cite{eiben2011tuning} and the irace package~\cite{lopezibanez2016irace} argue for explicit, budgeted tuning, and Bergstra and Bengio~\cite{bergstra2012random} show that random search is a strong baseline. We follow this view and tune with PSO against random search under a train/validation/test split.

%% file: sec_method.tex
\section{Problem Formulation}
\label{sec:problem}

\textbf{World and agents.} $N(t)$ drones with unique identifiers move in the square $[0,W]^2$. Drone $i$ has position $\mathbf p_i(t)$ and obeys the single-integrator model
\begin{equation}
\mathbf p_i(t{+}1)=\mathbf p_i(t)+\mathbf u_i(t)+\boldsymbol\eta_i(t),\qquad \|\boldsymbol\eta_i\|\le\varepsilon,
\label{eq:dyn}
\end{equation}
where $\mathbf u_i$ is the commanded displacement and $\boldsymbol\eta_i$ is actuation noise. Each noise coordinate is drawn from $\mathcal N(0,(m\sigma)^2)$ truncated at $\pm k\,m\sigma$ ($m$ is a difficulty multiplier), so that $\varepsilon=\sqrt2\,k\,m\sigma$. A \emph{collision} is any pair with $\|\mathbf p_i-\mathbf p_j\|<2r_d$ ($r_d$ is the drone radius).

\textbf{Sensing, communication and knowledge.} Every drone senses the relative positions of all drones within a short safety range $R$ (defined in Sec.~\ref{sec:bvc}); the safety filter uses only this information. In addition, every drone broadcasts one message per tick over a unit-disk radio graph of range $r_c\gg R$. The message contains its position and its claim table. Every directed message is lost independently with probability $q$, so a drone knows the positions of more distant drones only through the messages that reach it. A drone knows (i) the mission template, (ii) its own state and (iii) what its radio neighbours sent at the start of the tick. It is never told that a drone has been destroyed or has joined. In the \emph{partial-map} variant, slot $k$ is known to drone $i$ only while it lies within $r_s$ of the drone or of one of its current neighbours.

\textbf{Mission template.} A target picture is a set of $M$ slots $\{\mathbf s_k\}$ with priority ranks $\rho_k\in\{0,\dots,M{-}1\}$. The ranks come from farthest-point sampling of the shape (Sec.~\ref{sec:template}), so that the $K$ highest-priority slots form an evenly spread sub-sampling of the picture for \emph{every} $K\le M$ (Lemma~\ref{lem:fps}).

\textbf{Population changes.} At any tick an adversary may destroy any set of drones or release new drones from the launch pad. The swarm is never told.

\textbf{Metrics.} A slot is \emph{occupied} if a drone that has claimed it lies within $1.5\,r_h$ of it ($r_h$ is the hit radius). Let $K=\min(N,M)$.
\begin{itemize}
\item \emph{Formation time} $T_{95}=\min\{t:\ |\text{occupied}|/K\ge0.95\}$. A run \emph{succeeds} if $T_{95}$ is reached within the cap.
\item \emph{Priority fill} $\mathrm{PF}(t)$ is the fraction of the $K$ highest-priority slots that are occupied, i.e.\ how faithfully the picture is drawn after population changes. The \emph{excess Chamfer error} is the symmetric Chamfer distance between the drone positions and the full template, minus the same quantity for the ideal $K$-slot prefix.
\item \emph{Recovery}: after a disruption, $T_{\mathrm{rec},90}$ and $T_{\mathrm{rec},95}$ are the first ticks with $\mathrm{PF}\ge0.9$ and $\ge0.95$, and $\mathrm{PF}_{400}$ is the priority fill 400 ticks after the event. For $K=17$, $\mathrm{PF}\ge0.95$ requires all 17 slots, so we report the 90\% threshold as well.
\item \emph{Assignment ratio}: for the assignment when the run ends (at the time cap in Table~\ref{tab:x1full}, whose runs continue after formation, and at the moment of formation elsewhere), the sum over drones of the distance from the point where each drone claimed its current slot to that slot, divided by the optimal (Hungarian~\cite{kuhn1955hungarian,munkres1957}) matching cost from the same points to the \emph{same} set of occupied slots.
\item \emph{Safety}: the minimum pairwise distance over the whole run and the number of overlapping pair-ticks, which should ideally be zero.
\item \emph{Communication}: messages \emph{received} per drone per tick and claim-table entries held per drone. Each drone sends one broadcast per tick, so the first number is the mean number of radio neighbours heard.
\end{itemize}

\section{Method}
\label{sec:method}

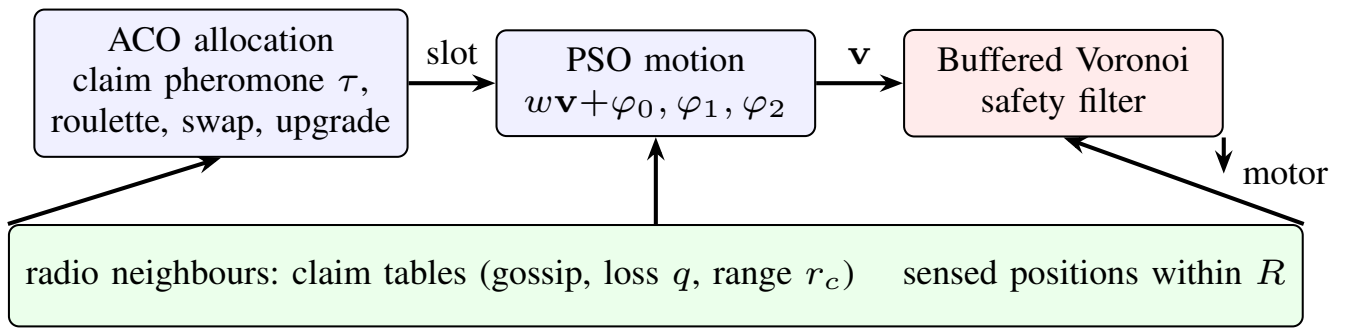
\begin{figure}[t]
\centering
\adjustbox{max width=\columnwidth}{\begin{tikzpicture}[font=\scriptsize,node distance=4mm,
 box/.style={draw,rounded corners=2pt,align=center,minimum height=7mm,minimum width=22mm,fill=blue!6},
 arr/.style={-{Stealth[length=1.6mm]},thick}]
\node[box] (aco) {ACO allocation\\claim pheromone $\tau$,\\roulette, swap, upgrade};
\node[box,right=6mm of aco] (pso) {PSO motion\\$w\mathbf v{+}\varphi_0,\varphi_1,\varphi_2$};
\node[box,right=6mm of pso,fill=red!8] (bvc) {Buffered Voronoi\\safety filter};
\node[box,below=6mm of pso,fill=green!8,minimum width=70mm] (radio) {radio neighbours: claim tables (gossip, loss $q$, range $r_c$) \quad sensed positions within $R$};
\draw[arr] (aco)--node[above]{slot}(pso);
\draw[arr] (pso)--node[above]{$\mathbf v$}(bvc);
\draw[arr] (bvc.south east)--++(0,-2.5mm)node[right]{motor};
\draw[arr] (radio.north west)--(aco.south);
\draw[arr] (radio.north)--(pso.south);
\draw[arr] (radio.north east)--(bvc.south);
\end{tikzpicture}}
\caption{Per-drone architecture. Nothing leaves the radio neighbourhood: there is no controller and no global state.}
\label{fig:arch}
\end{figure}

The controller of every drone has three layers with distinct roles (Fig.~\ref{fig:arch}). ACO decides \emph{which slot} to take, PSO decides \emph{how to fly} there, and a buffered-Voronoi filter decides \emph{how far it may safely move}. All three use only local information. A fourth, offline use of PSO tunes the hyper-parameters of the first two (Sec.~\ref{sec:tuning}).

\subsection{Design rationale}
\label{sec:rationale}
Each design choice answers one of the conditions listed in the introduction and replaces a simpler alternative that fails under it.

\textbf{Priority instead of a fixed assignment.} A fixed drone-to-slot plan needs the head-count, which no drone has. We instead order the slots so that every prefix is a good picture (Lemma~\ref{lem:fps}). Drones prefer important slots, losses are absorbed by the least important ones, and the question ``how many are we?'' never has to be answered.

\textbf{Pheromone instead of ownership messages.} An explicit protocol (``I take slot $k$'', ``I release slot $k$'', ``drone $j$ is dead'') breaks when messages are lost or a sender dies mid-protocol. A claim that must be continually reinforced, and otherwise evaporates, needs neither a release message nor a death notice: silence itself carries the information (Lemma~\ref{lem:ghost}). This is why we use the ACO \emph{mechanism} for allocation rather than an ACO-inspired search. An ablation that keeps every other component but never forgets isolates the value of evaporation (Sec.~\ref{sec:rq4}).

\textbf{One verdict per pair of claims.} When two drones compare the same pair of conflicting claims they must reach the same verdict; otherwise both may abandon the slot, or both may keep it. Ranking claims by reinforcement (with hysteresis) and then by distance makes the verdict a function of the pair alone (Lemma~\ref{lem:conflict}).

\textbf{PSO for motion, tested rather than assumed.} Under noise a pure proportional pull jitters, and a drone stuck behind others needs a way around. Inertia smooths the command, and following a neighbour that is closer to one's own goal opens a local lane through congestion. We keep PSO's personal-best and social terms but test their contribution separately.

\textbf{Safety outside the swarm-intelligence layers.} Allocation can be wrong for a while and parameters can be badly tuned; safety must depend on neither. The buffered Voronoi filter uses only sensed neighbour positions, is computed independently by each drone, and comes with a proof (Theorem~\ref{thm:safe}). The other layers can therefore be tuned and ablated freely without trading away safety.

\textbf{Offline PSO for the parameters.} Seventeen parameters interact, and the cost of a configuration is a noisy, non-differentiable average over stochastic simulations. Gradients are unavailable and a grid is infeasible in 17 dimensions, so a population-based black-box optimiser is the natural tool, with random search at the same budget as the control.

\subsection{Mission template}
\label{sec:template}
Shapes are defined as filled regions or stroke curves in $[-1,1]^2$. They are sampled by a hexagonal lattice (regions) or by arc-length spacing (strokes), with the spacing chosen so that at least $M$ points result. The points are ordered by farthest-point sampling (FPS), starting at the point nearest the centroid, and the first $M$ are kept; $\rho_k$ is the position in this order. A drone can therefore prefer low-rank slots without knowing how many drones are alive. We use four shapes (star, heart, letter D, smiley) for design and tuning, and eight further ones only to test generalisation (Sec.~\ref{sec:rq1}). The world side grows as $W=100\sqrt{\max(1,N/34)}$, so that the slot spacing stays above the safety distance.

\subsection{ACO: allocation by evaporating claim pheromone}
\label{sec:aco}
Each drone $i$ keeps a local table with one entry per slot: $(\text{owner},\ell,\text{stamp},\text{key})$. The \emph{claim pheromone} is
\begin{equation}
\tau_i(k,t)=\begin{cases}\ell\;2^{-(t-\text{stamp})/h}, & t-\text{stamp}\le\lceil 7h\rceil,\\ 0, & \text{otherwise},\end{cases}
\label{eq:tau}
\end{equation}
where $\ell\in[0,1]$ is the deposited level and $h$ the half-life (evaporation). An entry is \emph{fresh} if its age is at most $h$ and \emph{valid} if its age is at most $\lceil7h\rceil$.

\textbf{Deposit and reinforcement.} A drone holding slot $k$ refreshes its own entry every tick: $\ell\leftarrow\min(1,\ell+\delta)$, $\text{stamp}\leftarrow t$, $\text{key}\leftarrow\|\mathbf p_i-\mathbf s_k\|$. The longer a drone holds a slot, the stronger its claim, which is the positive-feedback half of ACO.

\textbf{Gossip.} Every tick drone $i$ merges into its own table the tables that its neighbours held at the start of the tick (double buffering, in the spirit of epidemic replication~\cite{demers1987epidemic}). Entry $b$ displaces entry $a$ for the same slot according to the following rule. If both have the same owner, the newer stamp wins. If one is fresh and the other is not, the fresh one wins. If both are fresh, the claim with the higher level wins when the levels differ by more than a hysteresis $\chi$; otherwise the smaller key (closer drone) wins, with ties broken by the smaller identifier. If neither is fresh, the newer stamp wins. After merging, a drone whose own slot is held by another owner's valid entry drops the slot (a \emph{conflict}). A voluntarily released slot is overwritten by a zero-level \emph{tombstone} with an infinite key, so stale copies of the old claim are not resurrected.

\textbf{Evaporation as failure detection.} Entries of destroyed drones are never refreshed, so $\tau\to0$ and the slot looks free again (Lemma~\ref{lem:ghost}). No message announces a death.

\textbf{Construction.} A drone without a slot builds one by roulette selection over the slots it knows (and has not given up on in the last 150 ticks):
\begin{equation}
P_i(k)\;\propto\;\big(1-\tau_i(k,t)\big)^{\alpha}\;\Big(\frac{1}{1+d_{ik}/d_0}\Big)^{\beta}\;e^{-\gamma\rho_k/M},
\label{eq:roulette}
\end{equation}
where $d_{ik}=\|\mathbf p_i-\mathbf s_k\|$. The first factor is the pheromone (a slot that others claimed recently is unattractive), the second the proximity heuristic and the third the priority preference. A drone that makes no progress towards its slot for $T_{\mathrm{stall}}$ ticks releases it and avoids it for 150 ticks.

\textbf{Local search.} Two unsettled neighbours exchange slots when this shortens the sum of their distances by more than $\mu$ and each is the other's best partner; mutual best choice makes the exchange consistent without a leader. A \emph{settled} drone (within $r_h$ of its slot) tries, with probability $\pi_{\mathrm{up}}$ per tick, to \emph{upgrade} to a slot whose rank is better by at least $\upsilon M$, whose pheromone is below $0.05$, and which lies within $r_{\mathrm{up}}=45$. Inside radio range the slot must be \emph{verifiably} empty, meaning no neighbour lies within $1.5r_h$ of it. Beyond radio range the slot must belong to the drone's known map and must carry no valid non-tombstone claim. Upgrades restore the high-priority picture after losses.

\textbf{Correspondence with ACO.} The layer keeps the ingredients of Ant System~\cite{dorigo1996antsystem}, with one deliberate change. Each drone is an ant, the slots are the solution components, and the claim table is the pheromone trail. Eq.~\eqref{eq:roulette} is the random-proportional transition rule, with the pheromone and heuristic weighted by the exponents $\alpha$ and $\beta$; Ant Colony System~\cite{dorigo1997acs} also deposits pheromone locally on the chosen component, as we do. Evaporation with half-life $h$ plays the role of the evaporation rate $\rho$. The change is the meaning of the pheromone. In ACO for optimisation, a deposit rewards solution quality and attracts ants. Here a deposit marks a component as taken and repels other ants, which enter Eq.~\eqref{eq:roulette} through $1-\tau$. Ant robots that cover terrain with evaporating, repelling traces~\cite{wagner1999evaporating,svennebring2004antrobots} use pheromone in the same coordinating way. We therefore regard the layer as ACO applied to distributed allocation rather than to search. Its ablation (Sec.~\ref{sec:rq4}) tests which of these ingredients matter.

\subsection{PSO: continuous motion}
\label{sec:pso}
A drone with slot $\mathbf s$ is a particle of a neighbourhood-best PSO with fitness
\begin{equation}
f_i(\mathbf p)=\|\mathbf p-\mathbf s\|+\lambda\sum_{j\in\mathcal N_i}\max(0,\,d_c-\|\mathbf p-\mathbf p_j\|),
\end{equation}
i.e.\ distance to the slot plus a crowding penalty ($d_c=4$). Its personal best $\mathbf b_i$ is the best position it has visited since taking the slot. Its neighbourhood best $\mathbf g_i$ is the position of the radio neighbour within $r_g$ that is closest to \emph{$i$'s} slot, if that neighbour is better than $i$ itself; this gives a leader to follow in congested flow. The velocity is
\begin{equation}
\begin{split}
\mathbf v_i\leftarrow{}& w\,\mathbf v_i+\varphi_0(\mathbf s-\mathbf p_i)\\&+\varphi_1 r_1(\mathbf b_i-\mathbf p_i)+\varphi_2 r_2(\mathbf g_i-\mathbf p_i),
\end{split}
\label{eq:pso}
\end{equation}
clamped to $\|\mathbf v_i\|\le v_{\max}$, with $r_1,r_2\sim U(0,1)$. The magnitude is clamped but not normalised, so the proportional term brakes smoothly near the slot (Proposition~\ref{prop:pso}). A drone that wants to move but is blocked for more than six consecutive ticks adds a lateral component of $0.8\,v_{\max}$ for 14 ticks, which breaks face-to-face deadlocks. In the partial-map variant, a drone that knows no free slot flies towards the template centroid and then explores with a persistent random heading until slots come into view.

Eq.~\eqref{eq:pso} is the inertia-weight PSO update~\cite{shi1998inertia}, with one difference: each particle also knows its own optimum, the slot, and is pulled towards it by $\varphi_0$. With $\varphi_1=\varphi_2=0$ it reduces to a damped proportional controller. Whether the stochastic personal- and social-best terms add anything to that controller is therefore an empirical question, and we test it both in normal swarms and under congestion (Sec.~\ref{sec:rq4}).

\subsection{Buffered Voronoi cells}
\label{sec:bvc}
Let $r_s=r_d+\varepsilon$ and $R=2v_{\max}+2r_s+2\varepsilon+1$. For every drone $j$ sensed within $R$, drone $i$ may only choose a next position $\mathbf q_i$ in the half-plane
\begin{equation}
\mathcal H_{ij}=\big\{\mathbf q:\ (\mathbf q-\mathbf m_{ij})\cdot\mathbf u_{ij}\ \ge\ r_s\big\},
\label{eq:bvc}
\end{equation}
where $\mathbf m_{ij}=(\mathbf p_i+\mathbf p_j)/2$ and $\mathbf u_{ij}=(\mathbf p_i-\mathbf p_j)/\|\mathbf p_i-\mathbf p_j\|$, and in the box $\mathcal B=[r_s,W-r_s]^2$. This is the buffered Voronoi cell of Zhou et al.~\cite{zhou2017bvc}, with the buffer enlarged by the noise bound. The desired point $\mathbf p_i+\mathbf v_i$ is projected onto $\mathcal C_i=\mathcal B\cap\bigcap_j\mathcal H_{ij}$ by cyclic projection onto the half-planes~\cite{bauschke1996projection}, with at most 200 sweeps. If the sweeps do not reach a point of $\mathcal C_i$ (in our experiments this happened in dense jams; see Sec.~\ref{sec:analysis} for when it can occur), the filter selects, among staying, the last iterate and a push-away step of length at most $v_{\max}$, the candidate with the largest worst-case constraint margin, and counts a \emph{compromise event}.

\subsection{Dynamic swarm size}
Nothing special is required. Destroyed drones stop refreshing their claims, which evaporate (Eq.~\ref{eq:tau}). Settled survivors then upgrade towards the freed high-priority slots, and drones without a slot (spares, or drones that lost a conflict) re-run the roulette. Joining drones start at a free launch-pad position (pad spacing $2r_s+0.4$) with an empty table. When $N>M$ the surplus drones wait as \emph{hot spares} and take over slots freed by later losses.

\subsection{PSO as a hyper-parameter optimiser}
\label{sec:tuning}
Seventeen parameters of the ACO and PSO layers (Table~\ref{tab:params}) are optimised offline. A candidate configuration is scored on 48 training runs: four shapes $\times$ three difficulties (normal, hard, brutal) $\times$ seeds 1--4. Each run forms the picture, waits 60 ticks, destroys 50\% of the drones and runs 400 more ticks. The cost of a run is
\begin{equation}
\begin{split}
J={}&T_{95}+0.5\,T_{\mathrm{rec},95}+300\,(1-\mathrm{PF}_{400})\\&+1000\cdot[\text{collision}]+0.1\,L/N,
\end{split}
\end{equation}
where failures are capped at 800 (formation) and 400 (recovery) ticks and $L$ is the total path length. The meta-optimiser is a standard global-best PSO (20 particles, 15 generations, inertia $0.72$, $c_1=c_2=1.49$~\cite{clerc2002constriction}), with one particle initialised at the hand-set values. Random search with the same budget (300 configurations, 14\,400 simulations) is the control~\cite{bergstra2012random}. Both finalists and the hand-set configuration are re-scored on 30 \emph{validation} seeds (500--529), and the better finalist is used. The meta-optimisers are themselves stochastic, so the whole procedure is run five times, with meta-seeds 2024 and 1--4. Two of the five configurations are used in the experiments: the \emph{reference} configuration (meta-seed 2024), on which all experiments on individual components are based, and the \emph{best-of-five} configuration, which has the lowest validation cost. Every number in Sec.~\ref{sec:results} uses \emph{test} seeds, which no optimiser ever saw: set~A (1001--1030) in all experiments, and set~B (2001--2030) in the formation comparison. For the generalisation study (Sec.~\ref{sec:rq4}) the same procedure is repeated on a \emph{narrow} training set: star and heart only, normal and hard only, eight seeds.

\section{Analysis}
\label{sec:analysis}

This section states what can be proved about the controller and where the proofs stop. Lemmas~\ref{lem:ghost}--\ref{lem:conflict} concern the ACO layer, Proposition~\ref{prop:pso} the PSO layer, and Theorem~\ref{thm:safe} the safety filter under arbitrary population changes.

\begin{lemma}[any prefix is a good picture]
\label{lem:fps}
Let $\mathcal P$ be the sampled shape points and $S_K$ the first $K$ points in FPS order. For every $K$, the covering radius $r(S_K)=\max_{\mathbf x\in\mathcal P}\min_{\mathbf s\in S_K}\|\mathbf x-\mathbf s\|$ is at most twice the smallest covering radius achievable by any $K$ points of $\mathcal P$.
\end{lemma}
\begin{proof}
This is Gonzalez's farthest-point guarantee for the $K$-center problem~\cite{gonzalez1985kcenter}. Let $\mathbf x^\ast$ be the point that FPS would pick next. Then the $K{+}1$ points $S_K\cup\{\mathbf x^\ast\}$ are pairwise at least $r(S_K)$ apart. Any $K$ centres must put two of these $K{+}1$ points in the same cluster, so some centre is at least $r(S_K)/2$ from one of them, by the triangle inequality. The proof holds for any starting point.
\end{proof}
\noindent Hence the swarm never needs to know $N$: whatever the number of survivors, filling the top-$K$ slots draws a picture that is uniformly covered up to a factor of two.

\begin{lemma}[ghost claims vanish everywhere]
\label{lem:ghost}
Let drone $a$ be destroyed at tick $t_a$. In every table of every surviving drone, at every tick $t\ge t_a$, every entry with owner $a$ has strength $\tau\le 2^{-(t-t_a)/h}$. It is not fresh after $t_a+h$ and not valid after $t_a+\lceil7h\rceil$. This holds for every topology, every loss pattern and every swarm size.
\end{lemma}
\begin{proof}
Only the owner writes a stamp: a refresh, construction, swap, upgrade or tombstone sets $\text{stamp}=t$ for its own entries. Gossip copies an entry verbatim, stamp included. Hence every copy of an entry owned by $a$ carries a stamp at most $t_a$, and its level is at most 1. Eq.~\eqref{eq:tau} gives the bound, and the freshness and validity thresholds are ages $h$ and $\lceil7h\rceil$.
\end{proof}
\noindent Lemma~\ref{lem:ghost} is the formal content of ``evaporation is failure detection''. A ghost claim can never block a slot for longer than $\lceil7h\rceil$ ticks, and after $h$ ticks any fresh claim of a living drone overrides it. The half-life $h$ thus trades failure-detection latency against tolerance to packet loss: a living holder whose messages are lost for more than $h$ consecutive ticks loses priority to a competitor.

\begin{lemma}[consistent resolution of pairwise conflicts]
\label{lem:conflict}
Let drones $a\ne b$ both hold slot $k$ with fresh entries $e_a,e_b$ at the start of tick $t$, and let no third claimant of $k$ be in their neighbourhoods. (i) The displacement rule is antisymmetric on $(e_a,e_b)$: exactly one entry wins. (ii) If on tick $t$ each drone receives the other's tick-start table, exactly one of them, the loser, releases $k$, so the conflict ends with neither a double release nor a double hold. (iii) If the two remain neighbours, such a two-way exchange occurs within $n$ ticks with probability at least $1-\big(1-(1-q)^2\big)^n$.
\end{lemma}
\begin{proof}
(i) If the levels differ by more than $\chi$, the higher level wins; otherwise the smaller key wins, and if the keys are equal, the smaller identifier. In each case exactly one entry wins, so $\text{displaces}(e_a,e_b)$ and $\text{displaces}(e_b,e_a)$ are never both true and never both false. (ii) Each drone's merged table starts from its own tick-start entry. It receives the other's tick-start entry directly, and that entry supersedes any older relayed copy of the same owner's claim (same owner, newer stamp). Both drones therefore compare exactly $e_a$ and $e_b$, and by (i) they reach the same verdict. (iii) Each of the two directed messages is lost independently with probability $q$ per tick.
\end{proof}
\noindent The lemma does not cover ticks on which only one direction is delivered. The drone that heard nothing then compares its claim with an older relayed copy of the other's claim, if it has one, and may reach a different verdict. A transient double release or double hold is then possible; it is undone at the next two-way exchange and is counted as a conflict. With three or more simultaneous claimants the rule is not transitive (because of the hysteresis), so the order of merging can matter. Conflicts are still detected, and the losers re-enter the roulette. We measure conflicts in every experiment.

\begin{proposition}[the PSO law converges to the slot]
\label{prop:pso}
Ignore noise, the safety filter, the clamp and the stochastic terms ($\varphi_1=\varphi_2=0$). Then the error $\mathbf e_t=\mathbf s-\mathbf p_t$ obeys a linear recursion with characteristic polynomial $\lambda^2-(1+w-\varphi_0)\lambda+w$. It converges to $0$ geometrically if and only if $|w|<1$ and $0<\varphi_0<2(1+w)$. If, in addition, $(1+w-\varphi_0)^2<4w$, the approach is an underdamped spiral with contraction factor $\sqrt{w}$ per tick.
\end{proposition}
\begin{proof}
From $\mathbf v_{t+1}=w\mathbf v_t+\varphi_0\mathbf e_t$ and $\mathbf e_{t+1}=\mathbf e_t-\mathbf v_{t+1}$, the state $(\mathbf v,\mathbf e)$ evolves by $\big[\begin{smallmatrix}w&\varphi_0\\-w&1-\varphi_0\end{smallmatrix}\big]$. Its trace is $1+w-\varphi_0$ and its determinant is $w$. The Jury conditions for a second-order polynomial, $|w|<1$ and $|1+w-\varphi_0|<1+w$, give the stated region. Complex roots have modulus $\sqrt{\det}=\sqrt{w}$. This is the classical deterministic PSO stability analysis~\cite{trelea2003pso,clerc2002constriction}, with the slot as the only attractor.
\end{proof}
\noindent Both the hand-set and the tuned values lie inside the region (Table~\ref{tab:params}). Far from the slot the clamp saturates and the drone flies at $v_{\max}$; near the slot the law takes over and brakes. The $\varphi_1,\varphi_2$ terms act as bounded perturbations whose attractors, the personal best and a neighbour position, move towards the slot as the drone progresses.

\begin{lemma}[safety for mutually sensed pairs]
\label{lem:pair}
If $\mathbf q_i\in\mathcal H_{ij}$, $\mathbf q_j\in\mathcal H_{ji}$ and $\|\boldsymbol\eta_i\|,\|\boldsymbol\eta_j\|\le\varepsilon$, then $\|(\mathbf q_i+\boldsymbol\eta_i)-(\mathbf q_j+\boldsymbol\eta_j)\|\ge2r_d$.
\end{lemma}
\begin{proof}
We have $\mathbf u_{ji}=-\mathbf u_{ij}$, and both half-planes use the same midpoint $\mathbf m$. So $(\mathbf q_i-\mathbf m)\!\cdot\!\mathbf u_{ij}\ge r_s$ and $(\mathbf q_j-\mathbf m)\!\cdot\!\mathbf u_{ij}\le-r_s$. Subtracting, $\|\mathbf q_i-\mathbf q_j\|\ge(\mathbf q_i-\mathbf q_j)\!\cdot\!\mathbf u_{ij}\ge2r_s$. The triangle inequality then gives distance at least $2r_s-2\varepsilon=2r_d$.
\end{proof}

\begin{theorem}[collision-freedom under arbitrary kills and joins]
\label{thm:safe}
Assume that (a) the drones present at $t=0$ and every drone that joins are at least $2r_d$ from all other drones when they appear; and (b) on every tick, the filter of every drone returns a point $\mathbf q_i\in\mathcal C_i$ with $\|\mathbf q_i-\mathbf p_i\|\le v_{\max}+\varepsilon$. Then $\|\mathbf p_i(t)-\mathbf p_j(t)\|\ge2r_d$ for all pairs and all ticks, for every schedule of destructions and joins and every swarm size.
\end{theorem}
\begin{proof}
By induction over ticks. Destructions only remove drones and constraints; joins are safe by (a). Consider a pair $i,j$ present at tick $t$ and $t{+}1$, and let $d=\|\mathbf p_i(t)-\mathbf p_j(t)\|$. If $d\le R$, each drone senses the other, so by (b) $\mathbf q_i\in\mathcal H_{ij}$ and $\mathbf q_j\in\mathcal H_{ji}$, and Lemma~\ref{lem:pair} applies. If $d>R$, then by (b) and the noise bound each drone moves at most $v_{\max}+2\varepsilon$, so the new distance is at least $R-2v_{\max}-4\varepsilon=2r_d+1$. Finally, $\mathbf q_i\in\mathcal B$ and $|\eta|\le\varepsilon/\sqrt2$ per coordinate keep every drone inside $[r_d,W-r_d]^2$, so the arena clamp never acts.
\end{proof}

\noindent\textbf{Where the guarantee stops.} Hypothesis (b) is automatic whenever $\mathbf p_i\in\mathcal C_i$, i.e.\ the drone's current position satisfies all of its own constraints. Then every projection is non-expansive towards $\mathbf p_i$, so all iterates stay within $v_{\max}$ of $\mathbf p_i$, and the sweeps converge to a point of $\mathcal C_i$~\cite{bauschke1996projection}. If the sweep budget runs out first, the ``stay'' candidate is itself feasible and the maximin rule selects a feasible point. Noise, however, can leave two drones at a distance in $[2r_d,2r_s)$. Then $\mathbf p_i\notin\mathcal H_{ij}$, and in a dense jam $\mathcal C_i$ can be empty. Theorem~\ref{thm:safe} then gives no guarantee, and safety becomes an empirical matter. For this reason we report, for every experiment, the ground-truth minimum separation over \emph{all} pairs and the number of overlapping pair-ticks. Positions are assumed to be sensed exactly; a sensing error bound would have to be added to $\varepsilon$.

\textbf{Cost.} Each drone broadcasts one table of $M$ entries per tick and receives one from each of its $|\mathcal N_i|$ radio neighbours. It therefore processes $O(|\mathcal N_i|\,M)$ entries per tick for gossip, plus $O(M)$ for the roulette. Neither term depends on $N$ explicitly. In our scaling experiments, however, $M=N$, and the number of neighbours also grew with $N$ (from 10 at $N=20$ and 17 at $N=34$ to 37 at $N=300$, averaged until the picture forms; Fig.~\ref{fig:x4}). The data received per drone thus grew faster than linearly in $N$. Averaged over a whole 1500-tick run, the number of neighbours at $N=34$ is lower (10.6, Table~\ref{tab:x1cost}). We do not model bandwidth limits; a protocol that gossips only recently changed entries would reduce this cost. The oracle needs global state and $O(N^3)$ Hungarian assignment at every population change.

%% file: sec_setup.tex
\section{Experimental Setup}
\label{sec:protocol}

\textbf{Simulator.} All results come from a deterministic, seeded simulator written in JavaScript; the same code drives the browser demonstration. Rerunning 1920 earlier runs reproduced every result exactly. One tick is one synchronous round of radio exchange, decision and motion. We use $r_d=1$, $v_{\max}=1$, noise $\sigma=0.05$ truncated at $2\sigma$ (so $\varepsilon=0.14$ at $m=1$) and hit radius $r_h=1$. Unless stated otherwise, the swarm has $N=34$ drones and the picture has $M=34$ slots. All quantities are in drone radii and ticks. As an illustration only, with a drone radius of 0.25\,m and a 10\,Hz control loop, $v_{\max}$ corresponds to 2.5\,m/s, the normal radio range to 6.25\,m, and a typical formation time of 100 ticks to 10\,s.

\textbf{Difficulty.} Four levels vary the radio range, motor noise, packet loss and how much of the picture a drone can see (Table~\ref{tab:presets}). The \emph{brutal} level combines a range of ten drone radii, 50\% message loss, three times the base noise and a partial map. In this regime the radio graph can break into disconnected groups.

\begin{table}[t]
\centering\footnotesize
\caption{Difficulty levels.}
\label{tab:presets}
\begin{tabular}{lrrrr}\toprule
Level & radio range $r_c$ & noise $m$ & packet loss $q$ & map radius $r_s$ \\\midrule
easy & 30 & 0.5 & 0 & $\infty$ \\
normal & 25 & 1 & 0.1 & $\infty$ \\
hard & 15 & 2 & 0.3 & 35 \\
brutal & 10 & 3 & 0.5 & 22 \\\bottomrule
\end{tabular}
\end{table}

\textbf{Compared controllers.} To attribute differences to allocation, all controllers use the same safety filter and, unless their own tuning includes motion, the motion parameters of our reference configuration. Our controller tunes its motion parameters together with the allocation parameters, and its hand-set variant keeps the hand-set ones.
\begin{itemize}
\item \emph{Central oracle}: optimal (Hungarian) assignment of all drones to the $K$ highest-priority slots with global knowledge, recomputed at every population change. No decentralised method can be expected to match it.
\item \emph{ACO-PSO (tuned)} and \emph{ACO-PSO (hand-set)}: the proposed controller with its reference configuration (Sec.~\ref{sec:tuning}) or with the hand-set parameters.
\item \emph{CBBA-like}: our tables, gossip and conflict rule, but argmax construction, claims that never expire, and no swaps or upgrades. It represents consensus without forgetting.
\item \emph{CBBA with reset}: the same, but claims time out when they are no longer refreshed, and agents re-bid for free slots of higher priority. This mirrors the reset and re-bidding that CBBA uses under failures~\cite{choi2009cbba}. Because it is built on our evaporating claim tables, it differs from our controller only in its construction rule (argmax instead of roulette) and in having no swaps. It is therefore both the strongest baseline and an ablation of ACO's stochastic construction. We report it with the hand-set allocation parameters and, for a fair comparison, tuned with the same budget and protocol as our controller over its ten allocation parameters (\emph{tuned}: reference configuration; \emph{best of five}). Its reference configuration is the random-search finalist of its tuning run, which had the lower validation cost (459.4 vs 469.7 for PSO). Table~\ref{tab:paramsel} lists the reference and best-of-five configurations of both tuned controllers.
\item \emph{Greedy nearest}: argmax with a strong proximity weight ($\alpha=4$, $\beta=8$, $\gamma=0$), with expiring claims and no swaps or upgrades.
\item \emph{Random slot}: a random slot among those believed free.
\end{itemize}

\textbf{Experiments.} Each research question is answered by dedicated experiments, all on held-out test seeds. The main comparisons between the two tuned controllers (formation, unseen shapes, repair after random losses, scale, radio range, map knowledge and spares) are made with both their reference and their best-of-five configurations; the full sweeps and the remaining repair scenarios use the reference configurations.
\begin{itemize}
\item \textbf{RQ1, formation.} A baseline comparison on four shapes and four difficulty levels with 30 seeds each, on both seed sets; all five tuned configurations of both tuned controllers; each construction rule run with the other controller's parameters; a scalability sweep from 20 to 300 drones, sweeps of packet loss, radio range, noise and map knowledge, and a test on eight shapes never used in design or tuning.
\item \textbf{RQ2, repair.} Random losses of 10--75\%, the loss of a whole region, cascades, joins, continuous churn, and losses in a swarm with more drones than slots (44 drones, 34 slots), where spares must take over.
\item \textbf{RQ3, safety.} The safety filter switched on and off at three noise levels, and a tally of every run of every experiment.
\item \textbf{RQ4, contributions.} One-at-a-time ablations, also repeated with 150 drones to test the PSO terms under congestion; a one-at-a-time parameter sensitivity sweep; PSO against random-search tuning, five runs per tuned controller with different meta-optimiser seeds; and a narrowly tuned configuration evaluated on unseen conditions.
\end{itemize}

\textbf{Statistics.} Seeds are identical across controllers, so all comparisons are paired. We report the following:
\begin{itemize}
\item success rates with Wilson 95\% intervals;
\item medians over successful runs with bootstrap 95\% intervals (2000 resamples);
\item PAR-2, a single score combining speed and reliability in which a failed run counts as twice the time cap;
\item paired Wilcoxon signed-rank tests on PAR-2;
\item for the five tuning repetitions, a paired $t$-test and a sign test on the validation costs;
\item exact McNemar or Fisher tests on success, and Mann--Whitney tests where configurations rather than runs are compared.
\end{itemize}
The time cap is 1500 ticks for formation (800--900 ticks when a disruption follows) and 400 ticks for recovery. All experiments ran on a single desktop CPU without a GPU. One script reproduces them (Sec.~\ref{sec:artefacts}).

%% file: sec_results.tex
\section{Results}
\label{sec:results}

\begin{figure*}[t]
\centering
\includegraphics[width=\textwidth]{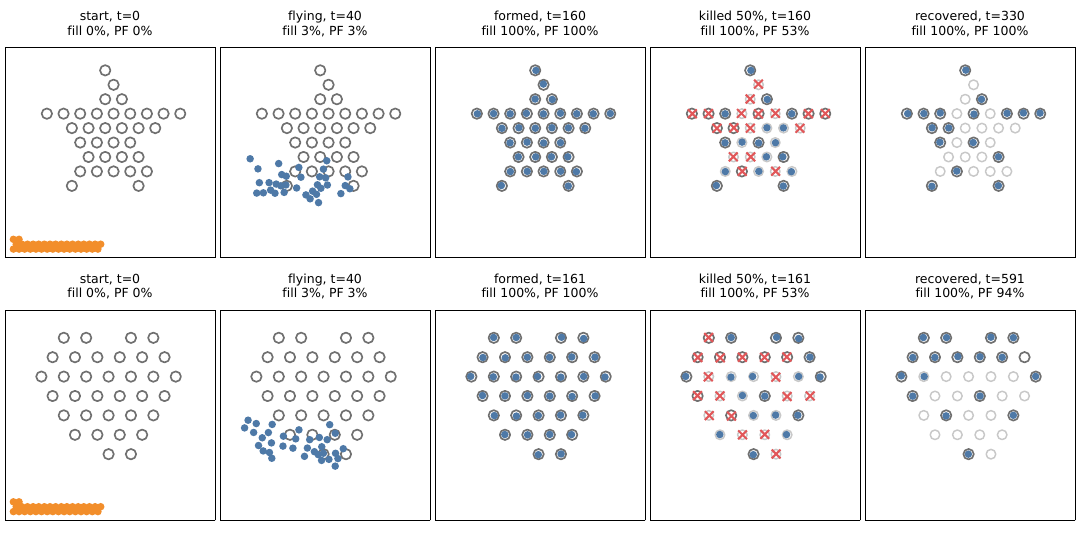}
\caption{A typical run (tuned controller, normal difficulty). Grey circles are slots (dark: the $K$ highest-priority ones), blue dots are drones, red crosses are destroyed drones. After half the swarm is destroyed without any notification, the survivors move from low- to high-priority slots, and the picture is redrawn at half resolution.}
\label{fig:snap}
\end{figure*}

All numbers below are from test seeds that no optimiser saw (Sec.~\ref{sec:tuning}). ``Tuned'' denotes the reference configuration of each tuned controller; the best-of-five configurations are reported alongside wherever the two tuned controllers are compared. Fig.~\ref{fig:snap} shows a typical run, and Appendix~\ref{app:extra} gives the full tables behind every summary.

\subsection{RQ1: Forming the picture}
\label{sec:rq1}

\textbf{Against other allocators.} Table~\ref{tab:x1} compares the controllers on four shapes with 30 seeds of set~A per difficulty level. In benign conditions every controller succeeds, and the comparison is about speed. The tuned controller forms the picture in a median of 104 ticks at normal difficulty, against 95 ticks for the oracle, and at the end of the run its assignments are within 1\% of the optimal matching. Allocators that do not forget fail as conditions worsen. CBBA-like fails in 16\% of hard runs and collapses to 5\% in the brutal regime: its claims never expire, so a claim that is wrong once blocks the slot for good. Random and greedy allocation succeed at hard difficulty only through assignments that, at the end of the run, are almost three times longer than optimal. In the brutal regime the tuned controller succeeds in 98\% of runs (95\% CI 93--99) with a median of 190 ticks. Greedy allocation reaches 89\% but takes three times as long, and the hand-set configuration reaches 72\%.

The most informative comparison is with CBBA with reset. This baseline uses the same evaporating claim tables, conflict rule and upgrades as our controller, and differs only in using argmax instead of roulette and in having no swaps. With hand-set parameters it reaches 89\% in the brutal regime. Tuned with the same budget, it matches our controller: 99\% success with a median of 183 ticks. Its PAR-2 is slightly lower at easy and normal difficulty (101 vs 104 and 102 vs 105, $p<0.001$), and the differences at hard and brutal difficulty are not significant ($p=0.32$ and $p=0.68$). On seed set~B (Table~\ref{tab:x14}) it was again marginally faster in benign conditions and more reliable in the brutal regime (99\% vs 92\% success, $p=0.005$), whereas our controller was marginally faster at hard difficulty ($p=0.02$). Its assignments were further from optimal than ours at hard and brutal difficulty. At the moment of formation (seed sets~A and~B) the assignment ratio was 1.13 vs 1.09 and 1.50 vs 1.33 with the reference configurations, and 1.25 vs 1.10 and 1.54 vs 1.32 with the best-of-five ones; at the end of the 1500-tick runs of Table~\ref{tab:x1full} it was 1.41 vs 1.08 and 1.49 vs 1.23. The baseline lacks the swaps, but we did not test whether they cause the difference.

The reference configurations happen to be the weakest of the five tuning runs for both controllers, with the highest validation cost (476.7 and 459.4, Table~\ref{tab:tunereps}) and the highest PAR-2 in the brutal regime (554 and 426, Table~\ref{tab:x16cfg}), so the results above are conservative for both. With the best-of-five configurations (Table~\ref{tab:x15}), both controllers succeeded in 99--100\% of the 240 runs per difficulty level, and the baseline was faster at every level (PAR-2 102 vs 133 at easy and 315 vs 367 at brutal difficulty; $p\le0.016$). Part of this gap is a selection effect: the lowest validation cost does not imply the fastest formation on test seeds, and our best-of-five configuration is one of our slower ones in benign conditions. Across all five configurations of each controller (Table~\ref{tab:x16cfg}), the baseline was faster at easy difficulty (in 24 of the 25 pairs of configurations; Mann--Whitney $p=0.02$) and marginally at normal difficulty ($p=0.06$), whereas at hard and brutal difficulty neither controller was consistently faster ($p=0.31$).

\textbf{Construction rule or parameters?} The tuned parameters of the two controllers differ (Table~\ref{tab:paramsel}), so a difference between them need not come from the construction rule. We therefore also ran each rule with the other controller's parameters, for the reference and for the best-of-five configurations (Table~\ref{tab:x16swap}). In the brutal regime, argmax without swaps was at least as reliable as roulette with swaps under all four parameter sets (99.2--99.6\% vs 90.8--99.2\% success), significantly so under three (McNemar $p\le0.013$). In benign conditions, by contrast, the rule that the parameters had been tuned for was always faster ($p<0.001$): with our best-of-five parameters argmax was 33\% slower at easy difficulty (PAR-2 177 vs 133), and with the baseline's best-of-five parameters roulette took twice as long (204 vs 102). The baseline's speed in benign conditions is therefore a property of its tuned parameters, whereas the reliability of deterministic construction under heavy loss held for every parameter set we tried. Table~\ref{tab:x16swap} changes the construction rule and the swaps together. The single-factor ablations of our controller (Table~\ref{tab:x2}) point to the rule: in the brutal regime, argmax alone raised success from 78\% to 84\% ($p=0.13$), whereas removing the swaps alone left it at 78\%. With equal tuning, the deterministic construction is at least as good as the stochastic one. The oracle remains the fastest everywhere, because it pays nothing for communication.

\begin{table}[t]
\centering\footnotesize
\caption{Formation from the launch pad: success rate (\%) and median ticks to 95\% of the picture (120 runs per cell). Full table with confidence intervals and assignment ratios: Table~\ref{tab:x1full}.}
\label{tab:x1}
\setlength{\tabcolsep}{3pt}
\adjustbox{max width=\linewidth}{\input{tables/x1_compact.tex}}
\end{table}

\begin{table}[t]
\centering\footnotesize
\caption{The baseline comparison on seed set~B (2001--2030; 120 runs per cell). PAR-2 combines speed and reliability. Every controller's PAR-2 differs significantly from that of ACO-PSO (tuned) at every difficulty (paired Wilcoxon $p<0.05$).}
\label{tab:x14}
\setlength{\tabcolsep}{2pt}
\adjustbox{max width=\linewidth}{\input{tables/x14.tex}}
\end{table}

\begin{table}[t]
\centering\footnotesize
\caption{Best-of-five configurations of both tuned controllers (lowest validation cost over five tuning runs, Table~\ref{tab:tunereps}) on seed sets~A and~B (240 runs per cell). The bottom row gives the paired Wilcoxon $p$ on PAR-2.}
\label{tab:x15}
\setlength{\tabcolsep}{2pt}
\adjustbox{max width=\linewidth}{\input{tables/x15.tex}}
\end{table}
\begin{table}[t]
\centering\footnotesize
\caption{Construction rule against parameters: each rule run with the tuned parameters of either controller (240 runs per cell, seed sets~A and~B). Roulette with swaps is our construction, argmax without swaps the baseline's. ``Paired $p$'': Wilcoxon on PAR-2 and, in the last column, McNemar on success, between the two rules under the same parameters.}
\label{tab:x16swap}
\setlength{\tabcolsep}{2pt}
\adjustbox{max width=\linewidth}{\input{tables/x16_swap.tex}}
\end{table}

\textbf{Beyond the training pictures.} Tuning used four shapes, so a fair question is whether the tuned parameters have merely learned those four. We applied the controllers, unchanged, to eight shapes never used in design or tuning: a disc, a ring, an arrow, a cross, two separate blobs, and three stroke shapes (letter A, a spiral and a sine wave). Table~\ref{tab:x9} shows that our controller formed every one of them in every run at normal and hard difficulty, and in 98\% of runs in the brutal regime; the only failures were on the disc. Here the tuned CBBA with reset generalised less well: it reached 88\% in the brutal regime (McNemar $p<0.001$), with 18 of its 19 failures on the disc, the one shape on which our controller also failed. The best-of-five configurations gave the same picture (98\% vs 89\%, $p<0.001$, Table~\ref{tab:x17}), again with all failures on the disc. CBBA-like fell to 2\%, and greedy matched our success rate but took more than twice as long.

\begin{table}[t]
\centering\footnotesize
\caption{Eight shapes never used in design or tuning (160 runs per cell). PF$_{400}$: priority fill 400 ticks after half of the swarm is destroyed. Per-shape results: Table~\ref{tab:x9shapes}.}
\label{tab:x9}
\setlength{\tabcolsep}{2.2pt}
\adjustbox{max width=\linewidth}{\input{tables/x9.tex}}
\end{table}

\begin{table}[t]
\centering\footnotesize
\caption{The two tuned controllers in specific regimes, with their reference and their best-of-five configurations (seed set~A). Repair succeeds when 90\% of the high-priority picture is restored within 400 ticks (80 runs; runs that had not formed before the loss are excluded); unseen shapes: 160 runs; 300 drones: 24 runs; radio range 8: 30 runs; map radius 15: 180 runs; spares (44 drones, 34 slots): priority fill 400 ticks after the loss, 80 runs. $p$: McNemar on success, Wilcoxon on ticks and priority fill.}
\label{tab:x17}
\setlength{\tabcolsep}{2pt}
\adjustbox{max width=\linewidth}{\input{tables/x17.tex}}
\end{table}

\textbf{Scale and environment.} Formation time grows sub-linearly with swarm size, as $N^{0.63}$, at the same rate as for the oracle (Fig.~\ref{fig:x4}, Table~\ref{tab:x4}). With 300 drones the reference configuration of our controller succeeded in all 24 runs, that of the tuned CBBA with reset in 16 ($p=0.008$) and CBBA-like in 13; with the best-of-five configurations both tuned controllers succeeded in all 24 runs (Table~\ref{tab:x17}). The robustness sweeps (Fig.~\ref{fig:x5}) show complementary weaknesses. Our controller tolerated up to 60\% packet loss at every radio range of 12 or more. At the shortest range (8) its reference configuration still succeeded in 77\% of runs at 80\% loss, whereas the tuned CBBA with reset reached 7\%, and only 50\% already at 60\% loss. The best-of-five configurations of both controllers were weaker at this range (53\% vs 50\% at 60\% loss, 50\% vs 0\% at 80\% loss). CBBA-like, whose claims never expire, still reached 97\%: when almost every message is lost, never forgetting is an advantage. Conversely, when a drone can see only 15 units of the picture, the reference configuration of our controller succeeded in 61\% of runs over all noise levels and that of the baseline in 89\%; the best-of-five configurations succeeded in 98--99\%. Reliability under such extreme conditions thus depends as much on the tuning run as on the construction rule. Each drone sends one message per tick, but the message carries one entry per slot, so bandwidth grows with the size of the picture. This is the main scalability cost of the design.

\subsection{RQ2: Repairing the picture}
\label{sec:rq2}

\begin{figure*}[t]
\centering
\includegraphics[width=0.85\textwidth]{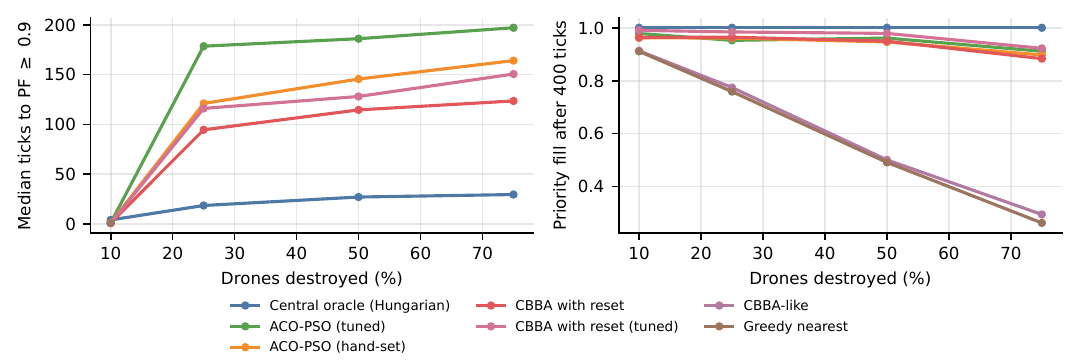}
\caption{Repair after silent losses (normal difficulty, 80 runs per point). Left: median ticks until 90\% of the high-priority picture is restored. Right: priority fill 400 ticks after the loss. Allocators without upgrades keep their slots occupied, but the picture degrades in proportion to the loss.}
\label{fig:x3}
\end{figure*}

Fig.~\ref{fig:x3} destroys a random 10--75\% of the drones after the picture has formed. Nobody is told. Each survivor only stops hearing the dead drones' claims, waits until they have faded, and then moves up to a vacated, more important slot. After half the swarm is lost, the tuned controller regains 90\% of the high-priority picture in 96\% of runs, with a median of 186 ticks and a priority fill of 0.96 after 400 ticks. After a 75\% loss it regains the picture in 71\% of runs. Allocators without upgrades also keep their slots occupied, but their picture is simply whichever half happened to survive (priority fill about 0.5).

Here the tuned CBBA with reset is better: it repairs faster (median 128 vs 186 ticks after a 50\% loss) and more reliably after a 75\% loss (88\% vs 71\% of runs, $p=0.01$). The best-of-five configurations show the same ranking (116 vs 188 ticks; 96\% vs 78\%, $p=0.001$; Table~\ref{tab:x17}). Losing a whole region, a cascade of losses and the arrival of new drones give the same ranking (Table~\ref{tab:x3}). Repair is also weaker for every controller at hard difficulty. On the unseen shapes, the priority fill 400 ticks after a 50\% loss was 0.85 for our controller, 0.82--0.87 for the CBBA variants and the hand-set configuration, and 1.00 for the oracle (Table~\ref{tab:x9}). When the swarm has spares (44 drones for 34 slots), our controller filled the gaps within a median of 37 ticks after a 25\% loss, faster than the oracle (64 ticks) and the tuned CBBA with reset (62 ticks), presumably because spares already wait near the picture (Table~\ref{tab:x11}). After a 50\% loss, however, the tuned CBBA with reset left a slightly fuller picture (priority fill 0.96 vs 0.93, and 0.98 vs 0.96 with the best-of-five configurations). Under continuous churn, with drones dying and joining throughout, our controller and both CBBA-with-reset variants kept 84--85\% priority fill, against 86\% for the oracle. Their geometric error was in fact lower than the oracle's, because the oracle re-solves the global assignment at every change and keeps many drones in transit.

Repair is several times slower than with the oracle (27 ticks after a 50\% loss), and the analysis explains much of the gap. A slot can be reused only after its ghost claim has faded below 0.05, which takes $h\log_2 20\approx79$ ticks for the tuned half-life (Lemma~\ref{lem:ghost}). This bound accounts for about 40\% of the median repair time; the remainder is spent waiting for an upgrade attempt and flying to the new slot. A shorter half-life would repair faster, but it would also mistake a drone whose messages are merely being lost for a dead one. The half-life therefore sets the trade-off between speed of repair and tolerance to packet loss. The hand-set configurations bear this out: with $h=12$ instead of the tuned 18, both our hand-set controller and the hand-set CBBA with reset repair faster but fail more often. The tuned CBBA with reset is an exception: with a similar half-life (19 ticks), it repairs faster. Our data do not isolate the reason; it upgrades about 2.5 times less often than our controller (15 vs 38 upgrades after a 50\% loss), which suggests fewer contested moves.

\subsection{RQ3: Avoiding collisions}
\label{sec:rq3}
\label{sec:rsafety}

Without the safety filter, every one of 720 runs ended with collisions (Table~\ref{tab:x7}), whatever the allocator and noise level. With the filter, no collision occurred in any of those runs. Table~\ref{tab:safety} pools every run of every experiment. Our controller, under every parameter set, produced overlaps in 5 of 21\,088 runs, in each case for one or two ticks and by at most 0.015 units (under 1\% of a diameter). In all five runs hypothesis (b) of Theorem~\ref{thm:safe} failed: the filter repeatedly met empty safety cells and had to fall back to the least-violating move. Four runs were jams in swarms of 150 or 300 drones, with thousands of such fallbacks; the oracle had two similar runs at $N=200$, so these overlaps come from the filter in jams rather than from the allocator. The fifth run had 34 drones and four times the base noise, so that $\varepsilon=0.57$. Noise of this size can leave two drones closer than $2r_s$, where a drone's own position violates its constraints; replaying the run showed a fallback on the very tick of the overlap. CBBA-like and the crippled ablation variants create such jams more often, because blocked or contested slots make many drones wait in the same place, and they show more and deeper overlaps (worst separation 1.66). The tuned CBBA with reset had shallow overlaps in 7 of 12\,004 runs (worst separation 1.97); the hand-set CBBA with reset and random allocation had none. The filter makes formation 14--23\% slower (median formation time, computed from the unrounded medians of Table~\ref{tab:x2}, last row). In short, it turns collisions from a certainty into rare, shallow events confined to jams. A filter that is safe even in jams remains an open problem.

\begin{table}[t]
\centering\footnotesize
\caption{All runs with the safety filter, from every experiment: overlaps (pairs closer than $2r_d$) by controller.}
\label{tab:safety}
\setlength{\tabcolsep}{3pt}
\adjustbox{max width=\linewidth}{\input{tables/safety.tex}}
\end{table}

\subsection{RQ4: What contributes, and does tuning generalise?}
\label{sec:rq4}
\label{sec:x8}
\label{sec:x9}

\textbf{Ablations.} Table~\ref{tab:x2} removes one component at a time. These runs use a formation cap of 800 ticks, because a 50\% loss follows formation, so brutal success rates are lower than in Table~\ref{tab:x1} (78\% vs 98\%): in Table~\ref{tab:x1}, 96 of the 120 brutal runs finish within 800 ticks and a further 21 between 800 and 1500 ticks, although the median is 190. Two components are indispensable in every regime. Without gossip no run ever forms the picture. Without evaporation formation takes nearly three times as long, only 24\% of brutal runs succeed, and repair after losses is poor (priority fill 0.65). Both are pheromone \emph{dynamics} of ACO. Upgrades are indispensable for repair: without them the picture after a 50\% loss stays at a priority fill of 0.50. Reinforcement was not ablated on its own, because without it every claim would stay at its initial level and the conflict rule would reduce to the distance key.

Other components are trade-offs rather than requirements. Resolving conflicts by distance instead of at random makes formation much faster at normal and hard difficulty (104 vs 179 and 137 vs 208 ticks, significant after Holm correction). Yet in the brutal regime random tie-breaking succeeded more often (92\% vs 78\%; McNemar $p<0.001$): under heavy loss, distance keys that change as drones move seem to prolong contested claims. Replacing the roulette, the textbook core of ACO construction, with a deterministic argmax costs 3\% at normal difficulty and does not hurt under stress (84\% vs 78\%, $p=0.13$). On the PSO side, the motion law matters only in the brutal regime, where a proportional controller lowers success from 78\% to 64\% ($p=0.001$). The personal-best, social and crowding terms have no measurable effect anywhere. We expected the social term to matter most in congestion, where following a neighbour could open a lane. With 150 drones, however, removing it, the crowding term or the whole PSO law changed the median formation time by less than 4\%, and none of the differences was significant ($p\ge0.07$, Table~\ref{tab:x12}). Upgrading into slots beyond radio range helps slightly at normal difficulty but hurts at hard difficulty, where the partial map makes ``no claim known'' a weak sign that a slot is free.

\begin{table}[t]
\centering\footnotesize
\caption{Ablations of the tuned controller (80 runs per cell, formation cap 800 ticks). $T_{95}$: median formation ticks; $^{*}$: $p<0.05$ against the full controller (paired Wilcoxon on PAR-2); $^{\dagger}$: still significant after Holm correction over the eleven variants. PF$_{400}$: priority fill 400 ticks after half of the swarm is destroyed.}
\label{tab:x2}
\setlength{\tabcolsep}{2.2pt}
\adjustbox{max width=\linewidth}{\input{tables/x2.tex}}
\end{table}

\begin{figure}[t]
\centering
\includegraphics[width=\columnwidth]{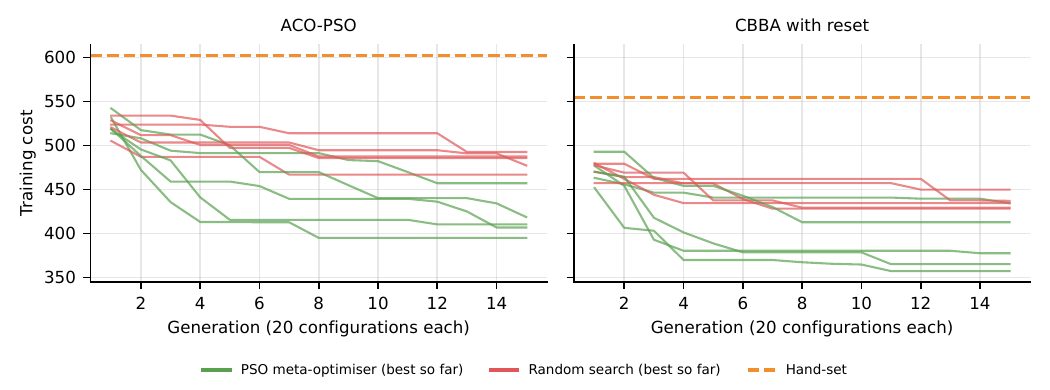}
\caption{Hyper-parameter tuning, five runs per controller with different meta-optimiser seeds: best training cost found so far by the PSO meta-optimiser and by random search (20 configurations per generation, 48 simulations each).}
\label{fig:tuning}
\end{figure}

\textbf{Tuning.} Both controllers were tuned five times with different meta-optimiser seeds, each time by PSO and by random search with the same budget of 300 configurations (Table~\ref{tab:tunereps}, Fig.~\ref{fig:tuning}). For our controller, PSO reached a better validation cost than random search in all five runs (mean 443.5 vs 487.2; paired $t$-test $p=0.012$, one-sided sign test $p=0.031$). For the baseline it was better in three of five runs (mean 419.7 vs 442.1, $p=0.17$). Over all ten runs PSO won eight times (sign test, one-sided $p=0.055$, two-sided $p=0.11$): it is the better tuner on average, but not in every run. Tuning mattered far more than the choice of optimiser: it lowered the validation cost from 594.4 to 429--477 for our controller and from 542.8 to 392--459 for the baseline. The configuration with the best validation cost is not necessarily the best on test seeds: our best-of-five configuration was more reliable than the reference configuration in the brutal regime (99\% vs 92--98\% success) but slower in benign conditions (median 128 vs 103 ticks at easy difficulty, Tables~\ref{tab:x1} and~\ref{tab:x15}), which is consistent with a tuning cost that contains no easy runs and weights repair heavily. The reference configuration's values are interpretable: strong avoidance of claimed slots, strong reinforcement and hysteresis (a held slot stays held), frequent upgrades and high inertia. Several values lie at the edge of the search range ($w$, $\mu$, $T_{\mathrm{stall}}$ and $\lambda$ at the maximum, $\upsilon$ at the minimum), so a wider range might improve the cost further. The one-at-a-time sweep (Fig.~\ref{fig:x6}) shows the cost still decreasing towards the edge for $\mu$, $T_{\mathrm{stall}}$ and $\upsilon$, and fluctuating within about 10\% for $w$ and $\lambda$. Both the tuned and the hand-set motion parameters lie inside the stability region of Proposition~\ref{prop:pso}. A one-at-a-time sweep shows that performance depends mostly on the pheromone exponent $\alpha$, the half-life and the upgrade rate; each motion parameter changes the cost by at most 15\% (Fig.~\ref{fig:x6}).

\begin{table}[t]
\centering\footnotesize
\caption{Validation cost (30 held-out seeds, lower is better) of the best configuration found by each meta-optimiser in five independent tuning runs per controller with the same budget. ``Selected'' is the better of the two.}
\label{tab:tunereps}
\adjustbox{max width=\linewidth}{\input{tables/tunereps.tex}}
\end{table}

\textbf{Generalisation of tuning.} To see what tuning can do wrong, we repeated it on a deliberately narrow training set of two shapes and two difficulty levels (Table~\ref{tab:x8}). On its own training conditions the narrow configuration beat the hand-set values by 16\%. Outside them it collapsed: in the brutal regime it formed the heart in no run (hand-set 63\%), letter D in 13\% (hand-set 90\%) and the smiley in 3\% (hand-set 33\%). The broadly tuned configuration was the fastest of the three in every cell and failed only in the hardest one (smiley, brutal, 90\%). Together with the unseen-shape results above, this shows that a tuned swarm is only as robust as its training distribution is broad. Tuning on a convenient subset can make it less robust than no tuning at all.

\begin{table}[t]
\centering\footnotesize
\caption{Generalisation of tuning: hand-set parameters, parameters tuned narrowly (star and heart, normal and hard only) and the broadly tuned parameters, on all shapes (30 runs per cell).}
\label{tab:x8}
\setlength{\tabcolsep}{2.5pt}
\adjustbox{max width=\linewidth}{\input{tables/x8.tex}}
\end{table}

%% file: tables/x1_compact.tex
\begin{tabular}{lrrrrrrrr}\toprule
 & \multicolumn{2}{c}{easy} & \multicolumn{2}{c}{normal} & \multicolumn{2}{c}{hard} & \multicolumn{2}{c}{brutal} \\
\cmidrule(lr){2-3}\cmidrule(lr){4-5}\cmidrule(lr){6-7}\cmidrule(lr){8-9}
Controller & succ. & $T_{95}$ & succ. & $T_{95}$ & succ. & $T_{95}$ & succ. & $T_{95}$ \\\midrule
Random slot & 100 & 109 & 100 & 113 & 100 & 168 & 62 & 684 \\
Greedy nearest & 100 & 116 & 100 & 118 & 100 & 280 & 89 & 583 \\
CBBA-like & 100 & 108 & 100 & 110 & 84 & 139 & 5 & 204 \\
CBBA with reset & 100 & 109 & 100 & 110 & 100 & 136 & 89 & 422 \\
CBBA with reset (tuned) & 100 & 101 & 100 & 104 & 100 & 128 & 99 & 183 \\
ACO-PSO (hand-set) & 100 & 118 & 100 & 122 & 100 & 148 & 72 & 631 \\
\textbf{ACO-PSO (tuned)} & 100 & 103 & 100 & 104 & 100 & 140 & 98 & 190 \\
\midrule
Central oracle & 100 & 94 & 100 & 95 & 100 & 96 & 100 & 100 \\
\bottomrule\end{tabular}

%% file: tables/x14.tex
\begin{tabular}{lrrrrrrrrrrrr}\toprule
 & \multicolumn{3}{c}{easy} & \multicolumn{3}{c}{normal} & \multicolumn{3}{c}{hard} & \multicolumn{3}{c}{brutal} \\
\cmidrule(lr){2-4}\cmidrule(lr){5-7}\cmidrule(lr){8-10}\cmidrule(lr){11-13}
Controller & succ. & $T_{95}$ & PAR-2 & succ. & $T_{95}$ & PAR-2 & succ. & $T_{95}$ & PAR-2 & succ. & $T_{95}$ & PAR-2 \\\midrule
Random slot & 100 & 110 & 110 & 100 & 110 & 111 & 100 & 170 & 228 & 65 & 858 & 1582 \\
Greedy nearest & 100 & 114 & 116 & 100 & 118 & 119 & 100 & 264 & 287 & 88 & 538 & 918 \\
CBBA-like & 100 & 108 & 109 & 100 & 110 & 110 & 87 & 142 & 530 & 4 & 182 & 2883 \\
CBBA with reset & 100 & 109 & 109 & 100 & 110 & 111 & 100 & 136 & 151 & 85 & 470 & 913 \\
CBBA with reset (tuned) & 100 & 100 & 100 & 100 & 104 & 103 & 100 & 131 & 137 & 99 & 203 & 425 \\
ACO-PSO (hand-set) & 100 & 118 & 121 & 100 & 124 & 124 & 100 & 148 & 158 & 72 & 628 & 1288 \\
\textbf{ACO-PSO (tuned)} & 100 & 104 & 103 & 100 & 106 & 105 & 100 & 132 & 133 & 92 & 188 & 620 \\
\midrule
Central oracle & 100 & 92 & 93 & 100 & 94 & 95 & 100 & 97 & 96 & 100 & 101 & 100 \\
\bottomrule\end{tabular}

%% file: tables/x15.tex
\begin{tabular}{lrrrrrrrrrrrr}\toprule
 & \multicolumn{3}{c}{easy} & \multicolumn{3}{c}{normal} & \multicolumn{3}{c}{hard} & \multicolumn{3}{c}{brutal} \\
\cmidrule(lr){2-4}\cmidrule(lr){5-7}\cmidrule(lr){8-10}\cmidrule(lr){11-13}
Controller & succ. & $T_{95}$ & PAR-2 & succ. & $T_{95}$ & PAR-2 & succ. & $T_{95}$ & PAR-2 & succ. & $T_{95}$ & PAR-2 \\\midrule
CBBA with reset (best of 5) & 100.0 & 103 & 102 & 100.0 & 105 & 104 & 100.0 & 142 & 147 & 99.6 & 197 & 315 \\
ACO-PSO (best of 5) & 100.0 & 128 & 133 & 100.0 & 123 & 132 & 100.0 & 153 & 154 & 99.2 & 213 & 367 \\
\midrule paired $p$ (PAR-2) & \multicolumn{3}{c}{$<$0.001} & \multicolumn{3}{c}{$<$0.001} & \multicolumn{3}{c}{0.007} & \multicolumn{3}{c}{0.016} \\
\bottomrule\end{tabular}

%% file: tables/x16_swap.tex
\begin{tabular}{lllrrrrr}\toprule
Configs & Parameters of & Construction & PAR-2 easy & normal & hard & brutal & brutal succ.\ \% \\\midrule
reference & ACO-PSO & roulette + swaps & 104 & 105 & 134 & 554 & 94.6 \\
 &  & argmax, no swaps & 109 & 112 & 137 & 439 & 99.2 \\
 & & \textit{paired $p$} & $<$0.001 & $<$0.001 & $<$0.001 & 0.005 & 0.003 \\
 & CBBA w.\ reset & roulette + swaps & 140 & 143 & 167 & 538 & 95.0 \\
 &  & argmax, no swaps & 101 & 103 & 136 & 426 & 99.2 \\
 & & \textit{paired $p$} & $<$0.001 & $<$0.001 & $<$0.001 & $<$0.001 & 0.013 \\
\midrule
best of five & ACO-PSO & roulette + swaps & 133 & 132 & 154 & 367 & 99.2 \\
 &  & argmax, no swaps & 177 & 182 & 205 & 292 & 99.6 \\
 & & \textit{paired $p$} & $<$0.001 & $<$0.001 & $<$0.001 & 0.653 & 1.000 \\
 & CBBA w.\ reset & roulette + swaps & 204 & 216 & 212 & 653 & 90.8 \\
 &  & argmax, no swaps & 102 & 104 & 147 & 315 & 99.6 \\
 & & \textit{paired $p$} & $<$0.001 & $<$0.001 & $<$0.001 & $<$0.001 & $<$0.001 \\
\bottomrule\end{tabular}

%% file: tables/x9.tex
\begin{tabular}{lrrrrrrrrr}\toprule
 & \multicolumn{3}{c}{normal} & \multicolumn{3}{c}{hard} & \multicolumn{3}{c}{brutal} \\
Method & Succ.\% & $T_{95}$ & PF$_{400}$ & Succ.\% & $T_{95}$ & PF$_{400}$ & Succ.\% & $T_{95}$ & PF$_{400}$ \\\midrule
Greedy nearest & 100 & 117 & -- & 100 & 322 & -- & 98 & 484 & -- \\
CBBA-like & 100 & 107 & -- & 93 & 138 & -- & 2 & 189 & -- \\
CBBA with reset & 100 & 108 & 0.93 & 100 & 139 & 0.86 & 76 & 394 & 0.79 \\
CBBA with reset (tuned) & 100 & 103 & 0.99 & 100 & 126 & 0.82 & 88 & 211 & 0.80 \\
ACO-PSO (hand-set) & 100 & 122 & 0.93 & 100 & 144 & 0.87 & 76 & 362 & 0.79 \\
ACO-PSO (tuned) & 100 & 103 & 0.97 & 100 & 125 & 0.85 & 98 & 222 & 0.80 \\
Central oracle (Hungarian) & 100 & 94 & 1.00 & 100 & 93 & 1.00 & 100 & 95 & 0.99 \\
\bottomrule\end{tabular}

%% file: tables/x17.tex
\begin{tabular}{lrrrrrr}\toprule
 & \multicolumn{3}{c}{reference configurations} & \multicolumn{3}{c}{best-of-five configurations} \\
\cmidrule(lr){2-4}\cmidrule(lr){5-7}
Test & ACO-PSO & CBBA-r & $p$ & ACO-PSO & CBBA-r & $p$ \\\midrule
Repair after losing 50\% (succ.\ \%) & 96 & 96 & 1.000 & 100 & 100 & 1.000 \\
\quad median ticks & 186 & 128 & $<$0.001 & 188 & 116 & $<$0.001 \\
Repair after losing 75\% (succ.\ \%) & 71 & 88 & 0.011 & 78 & 96 & 0.001 \\
\quad median ticks & 197 & 150 & $<$0.001 & 196 & 128 & $<$0.001 \\
Unseen shapes, hard (succ.\ \%) & 100 & 100 & 1.000 & 100 & 100 & 1.000 \\
Unseen shapes, brutal (succ.\ \%) & 98 & 88 & $<$0.001 & 98 & 89 & $<$0.001 \\
300 drones (succ.\ \%) & 100 & 67 & 0.008 & 100 & 100 & 1.000 \\
Radio range 8, 60\% loss (succ.\ \%) & 97 & 50 & $<$0.001 & 53 & 50 & 1.000 \\
Radio range 8, 80\% loss (succ.\ \%) & 77 & 7 & $<$0.001 & 50 & 0 & $<$0.001 \\
Map radius 15, all noise levels (succ.\ \%) & 61 & 89 & $<$0.001 & 99 & 98 & 0.688 \\
Spares, losing 50\% (PF after 400 ticks) & 0.93 & 0.96 & $<$0.001 & 0.96 & 0.98 & 0.003 \\
\bottomrule\end{tabular}

%% file: tables/safety.tex
\begin{tabular}{lrrrr}\toprule
Controller (safety layer on) & runs & runs with overlap & overlap pair-ticks & min sep \\\midrule
ACO-PSO (all parameter sets) & 21088 & 5 & 6 & 1.99 \\
CBBA-like & 4262 & 46 & 108 & 1.66 \\
CBBA with reset & 2720 & 0 & 0 & 2.02 \\
CBBA with reset (tuned) & 12004 & 7 & 7 & 1.97 \\
Greedy nearest & 2080 & 1 & 1 & 2.00 \\
Central oracle (Hungarian) & 4742 & 2 & 2 & 1.96 \\
Random slot & 1140 & 0 & 0 & 2.03 \\
ablated variants & 8712 & 130 & 172 & 1.75 \\
\bottomrule\end{tabular}

%% file: tables/x2.tex
\begin{tabular}{lrrrrrrrrr}\toprule
 & \multicolumn{3}{c}{normal} & \multicolumn{3}{c}{hard} & \multicolumn{3}{c}{brutal} \\
Variant (tuned) & Succ.\% & $T_{95}$ & PF$_{400}$ & Succ.\% & $T_{95}$ & PF$_{400}$ & Succ.\% & $T_{95}$ & PF$_{400}$ \\\midrule
full controller & 100 & 104 & 0.96 & 100 & 137 & 0.79 & 78 & 166 & 0.81 \\
without swaps & 100 & 114$^{\dagger}$ & 0.98 & 100 & 134$^{*}$ & 0.77 & 78 & 178$^{*}$ & 0.83 \\
without upgrades & 100 & 105 & 0.50 & 100 & 144$^{\dagger}$ & 0.60 & 72 & 410$^{\dagger}$ & 0.62 \\
upgrades within radio range only & 100 & 104 & 0.94 & 100 & 146$^{\dagger}$ & 0.88 & 75 & 192$^{\dagger}$ & 0.83 \\
without gossip & 0 & -- & -- & 0 & -- & -- & 0 & -- & -- \\
without evaporation & 99 & 283$^{\dagger}$ & 0.65 & 86 & 322$^{\dagger}$ & 0.71 & 24 & 383$^{\dagger}$ & 0.73 \\
random conflict tie-break & 99 & 179$^{\dagger}$ & 0.89 & 100 & 208$^{\dagger}$ & 0.73 & 92 & 226 & 0.76 \\
argmax instead of roulette & 100 & 107$^{\dagger}$ & 0.97 & 100 & 132 & 0.86 & 84 & 163 & 0.84 \\
proportional motion (no PSO) & 100 & 104 & 0.95 & 100 & 138 & 0.79 & 64 & 196$^{\dagger}$ & 0.84 \\
without personal/social terms & 100 & 104 & 0.97 & 100 & 135 & 0.80 & 76 & 160 & 0.82 \\
without crowding term & 100 & 103 & 0.98 & 100 & 138 & 0.82 & 80 & 164 & 0.82 \\
without safety filter & 100 & 89$^{\dagger}$ & 0.95 & 100 & 112$^{\dagger}$ & 0.79 & 81 & 145$^{\dagger}$ & 0.81 \\
\bottomrule\end{tabular}

%% file: tables/tunereps.tex
\begin{tabular}{lrrrrrr}\toprule
 & \multicolumn{3}{c}{ACO-PSO} & \multicolumn{3}{c}{CBBA with reset} \\
\cmidrule(lr){2-4}\cmidrule(lr){5-7}
Meta-seed & PSO & RS & selected & PSO & RS & selected \\\midrule
2024 & 476.7 & 507.5 & 476.7 & 469.7 & 459.4 & 459.4 \\
1 & 429.9 & 509.2 & 429.9 & 391.7 & 426.0 & 391.7 \\
2 & 428.7 & 480.4 & 428.7 & 393.1 & 439.1 & 393.1 \\
3 & 442.9 & 472.1 & 442.9 & 446.5 & 438.1 & 438.1 \\
4 & 439.4 & 466.7 & 439.4 & 397.4 & 447.9 & 397.4 \\
\midrule mean $\pm$ sd & 443.5 $\pm$ 19.5 & 487.2 $\pm$ 19.9 & 443.5 $\pm$ 19.5 & 419.7 $\pm$ 36.1 & 442.1 $\pm$ 12.4 & 415.9 $\pm$ 30.9 \\
hand-set & \multicolumn{3}{c}{594.4} & \multicolumn{3}{c}{542.8} \\
\bottomrule\end{tabular}

%% file: tables/x8.tex
\begin{tabular}{llrrrrrr}\toprule
 & & \multicolumn{2}{c}{normal} & \multicolumn{2}{c}{hard} & \multicolumn{2}{c}{brutal} \\
Shape & Parameters & Succ.\% & Median $T_{95}$ & Succ.\% & Median $T_{95}$ & Succ.\% & Median $T_{95}$ \\\midrule
star (train) & hand-set & 100 & 114 & 100 & 130 & 100 & 166 \\
 & tuned-narrow & 100 & 106 & 100 & 124 & 100 & 170 \\
 & tuned-broad & 100 & 102 & 100 & 111 & 100 & 144 \\
\midrule
heart (train) & hand-set & 100 & 115 & 100 & 128 & 63 & 452 \\
 & tuned-narrow & 100 & 104 & 100 & 124 & 0 & -- \\
 & tuned-broad & 100 & 101 & 100 & 112 & 100 & 163 \\
\midrule
letterD (unseen) & hand-set & 100 & 126 & 100 & 163 & 90 & 930 \\
 & tuned-narrow & 100 & 116 & 100 & 180 & 13 & 360 \\
 & tuned-broad & 100 & 104 & 100 & 147 & 100 & 312 \\
\midrule
smiley (unseen) & hand-set & 100 & 140 & 100 & 207 & 33 & 1284 \\
 & tuned-narrow & 100 & 124 & 90 & 444 & 3 & 1383 \\
 & tuned-broad & 100 & 111 & 100 & 160 & 90 & 1066 \\
\midrule
\bottomrule\end{tabular}

%% file: sec_discussion.tex
\section{Discussion}
\label{sec:discussion}

\textbf{What the results say about ACO and PSO.} The experiments give a more specific answer than ``swarm intelligence helps''. What carries the allocation is not ACO's probabilistic construction but its pheromone \emph{dynamics}: claims that are reinforced while held, evaporate when silent, and spread by gossip. Evaporation turns silence into failure detection at no communication cost (Lemma~\ref{lem:ghost}). Remove evaporation or gossip and the swarm breaks. Replace the roulette with argmax and drop the swaps, as the CBBA-with-reset baseline does, and the controller does not get worse: tuned with the same budget, both formed the picture in at least 94\% of runs at every difficulty level (seed sets A and B pooled), the deterministic rule was at least as reliable under heavy loss with every parameter set we tried, and in benign conditions which of the two was faster depended on the parameters rather than on the rule. An allocator that never forgets (CBBA-like) collapses under loss, and allocators that ignore the priority order and never upgrade (greedy, random) form slowly in hard conditions and cannot repair the picture.

On the PSO side, inertia and a pull towards the goal are needed in the hardest regime, but the personal-best and social terms contributed nothing we could measure, not even in a congested swarm of 150 drones. PSO's clearest contribution was as an offline optimiser. Tuning, rather than the construction rule, separated the strong configurations from the weak ones: it lowered the validation cost of both controllers by 15--28\%. PSO beat random search in all five runs for our controller and in three of five for the baseline, so it is the better tuner on average but not in every run. This suggests a division of labour for physical swarms: evaporating pheromone claims for discrete coordination, and a population-based optimiser such as PSO for the parameters.

\textbf{Trade-offs.} The two tuned allocators fail in different ways. With both of its configurations, our controller generalised better to unseen shapes in the brutal regime, kept its assignments closer to optimal, and was more reliable at the shortest radio range under 80\% packet loss, although its best-of-five configuration was much weaker there than its reference one. With both of its configurations, the tuned CBBA with reset repaired faster and more reliably after large losses. Other differences, with 300 drones and with a small map radius, disappeared with the best-of-five configurations; they belong to the tuning run rather than to the construction rule. Both allocators share three costs. First, repair is limited by evaporation: a slot cannot be reused before its ghost claim fades, and a shorter half-life would mistake drones with lossy links for dead ones. Second, the safety filter makes formation 14--23\% slower. The oracle, which pays nothing for communication, stays faster in formation and in most repair scenarios; only spare drones waiting near the picture were quicker to fill gaps than its global reassignment. Third, an optimiser exploits every metric it is not asked to care about. The inertia of the reference configuration, $w=0.9$, is stable (Proposition~\ref{prop:pso}) but integrates noise. Tuned drones flew almost the same distance to reach the picture as hand-set ones (71.2 vs 68.5 units per drone). While hovering, however, they moved twice as much (0.156 vs 0.075 units per tick), and this station keeping dominates the path lengths in Table~\ref{tab:x1cost}. The cause is that the tuning objective weighted hovering energy only briefly; lowering the inertia once a drone has settled would remove the effect. The narrow-tuning experiment shows the same effect at the level of scenarios.

\textbf{Threats to validity.}
\label{sec:threats}
\begin{itemize}
\item \emph{Test seeds and one design revision.} One rule was revised after results on test set~A had been inspected. The earlier version treated a slot as occupied whenever any neighbour was within three units of it. On compact stroke shapes neighbouring slots are closer than that, so survivors refused vacated slots and repair stalled at about 88\% priority fill. The corrected rule was checked on validation seeds, and tuning and all experiments use it. Set~A has therefore influenced one design decision; set~B has not. On set~B the ranking against the untuned baselines is the same, but our controller's success in the brutal regime was 92\% rather than 98\%, and the tuned CBBA with reset was significantly better in that regime (Table~\ref{tab:x14}), which indicates how much such comparisons vary from one seed set to another.
\item \emph{Simulation only.} We model 2D single-integrator dynamics with synchronous rounds and exact sensing of nearby positions. Wind, batteries, 3D flight, asynchronous radio and real quadrotor dynamics are not represented. Sensing error would have to be added to the noise bound of Theorem~\ref{thm:safe}.
\item \emph{Baselines.} Our CBBA-style baselines reproduce the single-assignment core of CBBA, with and without reset, inside our own message format rather than as the full bundle algorithm. Greedy, random and CBBA-like are not tuned. The oracle reassigns instantly and at no cost.
\item \emph{Tuning range.} Several tuned values lie at the edge of the search range, so better configurations may exist for both tuned controllers.
\item \emph{Scope.} Twelve 2D shapes were tested. Templates are known in advance and are static, and shapes whose slots would lie closer than the safety distance cannot be drawn. Surplus drones only wait as spares.
\end{itemize}

\section{Conclusion}
We asked whether a drone swarm can form and maintain a picture without a coordinator under the conditions that make this hard: an unknown head-count, silent failures, lossy radio, noisy motion and sensitive parameters. Our answer combines ACO pheromone dynamics for deciding who goes where, a PSO law for flying, a buffered Voronoi filter for separation, and PSO for offline tuning. We proved four things: claims of destroyed drones disappear within bounded time; a pairwise conflict is resolved consistently when the two drones exchange tables; any number of survivors covers the sampled picture to within a factor of two of the best possible; and collisions are impossible while safety cells are feasible. We also showed that the motion law is stable.

In experiments on held-out seeds, the controller formed the picture in 95\% of 240 runs in the hardest regime with its reference configuration (98\% and 92\% on the two seed sets) and in 99\% with its best-of-five configuration, repaired it after the silent loss of half the swarm, transferred unchanged to unseen shapes, scaled to 300 drones, and overlapped in only 5 of 21\,088 runs. Allocators that never forget, or that ignore the priority order and never upgrade, fell far behind. A baseline that shares those claims but constructs deterministically was, once tuned with the same budget, at least as reliable as our controller and repaired faster, while ours generalised better to unseen shapes. Run with identical parameters, the deterministic rule was at least as reliable under heavy loss, and in benign conditions speed depended on the parameters rather than on the rule. The evidence therefore points to the pheromone dynamics and to parameter tuning, rather than to ACO's stochastic construction or PSO's social terms, as what makes such swarms work. The controller remains slower than a centralised oracle, and its repair speed is bounded by evaporation. Natural next steps are a safety filter that is provably safe in jams, asynchronous 3D models, and flights on real drones.

%% file: sec_appendix.tex
\appendix

\section{Parameters}
\label{app:params}
Table~\ref{tab:params} lists the seventeen tuned parameters. Fixed constants: $r_d=1$, $v_{\max}=1$, $\sigma=0.05$, truncation $k=2$, $r_h=1$, crowding distance $d_c=4$, upgrade radius $r_{\mathrm{up}}=45$, stall cool-down 150 ticks, launch-pad spacing $2r_s+0.4$. Parameter names map to the symbols of Sec.~\ref{sec:method}: \texttt{alpha}~$\alpha$, \texttt{beta}~$\beta$, \texttt{gamma}~$\gamma$, \texttt{halfLife}~$h$, \texttt{deposit}~$\delta$, \texttt{dscale}~$d_0$, \texttt{swapMargin}~$\mu$, \texttt{upMargin}~$\upsilon$, \texttt{upRate}~$\pi_{\mathrm{up}}$, \texttt{stallTicks}~$T_{\mathrm{stall}}$, \texttt{w}~$w$, \texttt{phi0..2}~$\varphi_{0..2}$, \texttt{socR}~$r_g$, \texttt{lam}~$\lambda$, \texttt{hysteresis}~$\chi$.

\begin{table}[h]
\centering\scriptsize
\caption{Parameters of our controller: search range, hand-set value, and the PSO and random-search finalists of the reference tuning run (meta-seed 2024). The PSO finalist had the lower validation cost and is the reference configuration.}
\label{tab:params}
\adjustbox{max width=\linewidth}{\input{tables/params.tex}}
\end{table}

\begin{table}[h]
\centering\scriptsize
\caption{The configurations used in the experiments: reference (meta-seed 2024) and best of five (lowest validation cost) of both tuned controllers. The baseline has no swaps and therefore no swap margin, and its tuning leaves the motion parameters ($w$ to $\lambda$) at the values of our reference configuration. Table~\ref{tab:x16swap} runs each construction rule with each of these parameter sets; when our rule runs with the baseline's parameters, the swap margin keeps its hand-set value.}
\label{tab:paramsel}
\adjustbox{max width=\linewidth}{\input{tables/params_sel.tex}}
\end{table}

\begin{table}[h]
\centering\scriptsize
\caption{All five tuned configurations of each controller on the seeds of Tables~\ref{tab:x1} and~\ref{tab:x14} (240 runs per cell). Validation cost and the meta-optimiser whose finalist was selected; ref.: reference configuration; $^{b}$: best of five. Last row: in how many of the 25 pairs of configurations ACO-PSO had the lower PAR-2, and the Mann--Whitney $p$ over the five PAR-2 values per controller.}
\label{tab:x16cfg}
\setlength{\tabcolsep}{2pt}
\adjustbox{max width=\linewidth}{\input{tables/x16_configs.tex}}
\end{table}

\section{Supplementary results}
\label{app:extra}

\begin{table*}[t]
\centering\scriptsize
\caption{Formation from the launch pad, full results (120 runs per cell). Success rate with Wilson 95\% CI; median ticks over successful runs with bootstrap 95\% CI; PAR-2 (failures count as 3000 ticks); assignment ratio at the end of the run, tick 1500 (1 = optimal); paired Wilcoxon $p$ on PAR-2 against ACO-PSO (tuned).}
\label{tab:x1full}
\adjustbox{max width=\linewidth}{\input{tables/x1.tex}}
\end{table*}

\begin{table*}[t]
\centering\scriptsize
\caption{Repair at normal difficulty (80 runs per cell). Each cell: median ticks until 90\% priority fill / \% of runs reaching it within 400 ticks / mean priority fill after 400 ticks. Bottom: continuous churn.}
\label{tab:x3}
\adjustbox{max width=\linewidth}{\input{tables/x3.tex}}
\end{table*}

\begin{table}[t]
\centering\scriptsize
\caption{Communication and effort per run (normal difficulty). Path length covers the full 1500-tick run and is dominated by station keeping (Sec.~\ref{sec:discussion}).}
\label{tab:x1cost}
\setlength{\tabcolsep}{3pt}
\adjustbox{max width=\linewidth}{\input{tables/x1_cost.tex}}
\end{table}

\begin{table}[t]
\centering\scriptsize
\caption{Safety filter on and off (60 runs per row and setting; noise as a multiple of the base level).}
\label{tab:x7}
\setlength{\tabcolsep}{2.4pt}
\adjustbox{max width=\linewidth}{\input{tables/x7.tex}}
\end{table}

\begin{table}[t]
\centering\scriptsize
\caption{Spare drones: 44 drones for 34 slots, normal difficulty (80 runs per cell). Median ticks until 90\% priority fill (and \% of runs reaching it within 400 ticks), and priority fill after 400 ticks.}
\label{tab:x11}
\setlength{\tabcolsep}{3pt}
\adjustbox{max width=\linewidth}{\input{tables/x11.tex}}
\end{table}

\begin{table}[t]
\centering\scriptsize
\caption{PSO terms under congestion: 150 drones, normal difficulty, star and smiley (24 runs per row).}
\label{tab:x12}
\setlength{\tabcolsep}{3pt}
\adjustbox{max width=\linewidth}{\input{tables/x12.tex}}
\end{table}

\begin{table}[t]
\centering\scriptsize
\caption{Scalability: log--log fit of formation time against swarm size (bootstrap 95\% CI).}
\label{tab:x4}
\adjustbox{max width=\linewidth}{\input{tables/x4.tex}}
\end{table}

\begin{table}[t]
\centering\scriptsize
\caption{Formation success (\%) on each unseen shape (20 runs per cell).}
\label{tab:x9shapes}
\setlength{\tabcolsep}{2.4pt}
\adjustbox{max width=\linewidth}{\input{tables/x9_shapes.tex}}
\end{table}

\begin{figure*}[t]
\centering
\includegraphics[width=\textwidth]{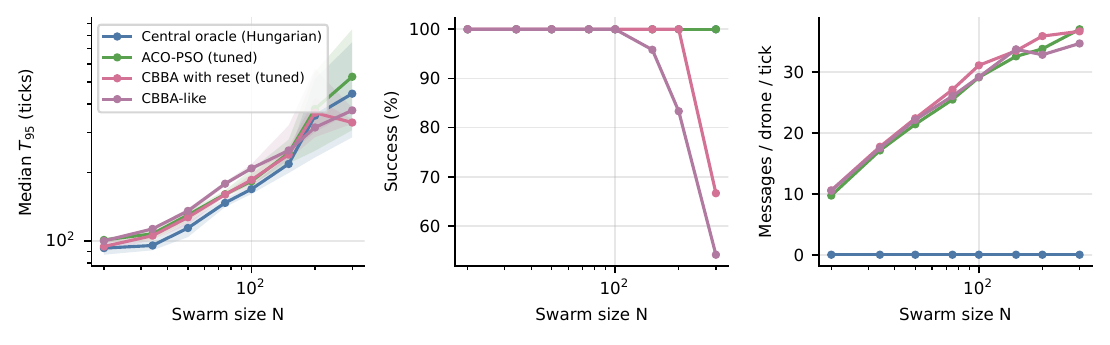}
\caption{Scalability (star and smiley, 12 seeds per size; world side $\propto\sqrt N$): formation time, success rate and messages per drone per tick.}
\label{fig:x4}
\end{figure*}

\begin{figure*}[t]
\centering
\includegraphics[width=.49\textwidth]{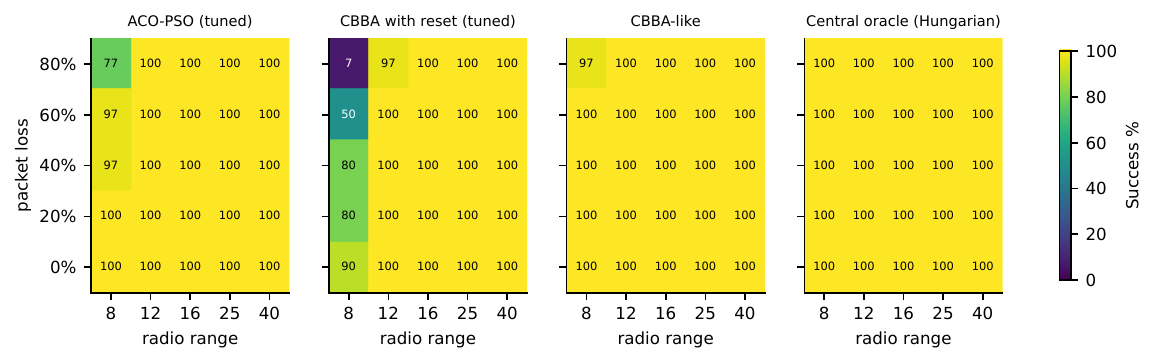}\hfill
\includegraphics[width=.49\textwidth]{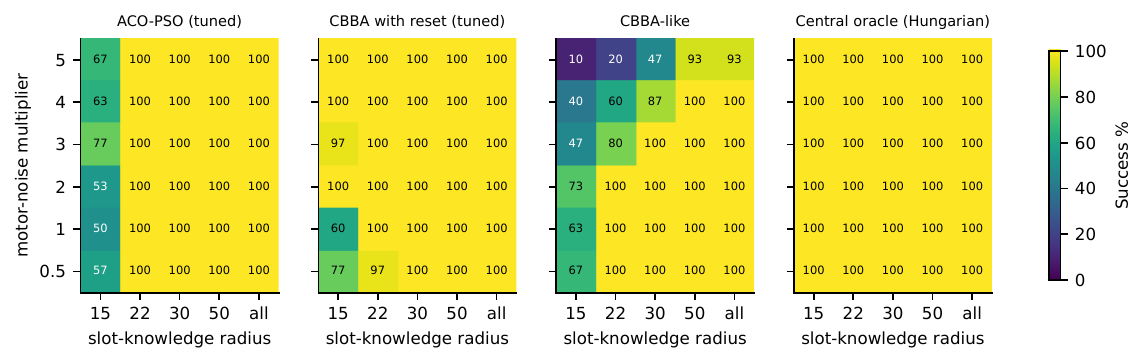}
\caption{Robustness (star and smiley, 15 seeds per cell): success rate (\%) against packet loss and radio range (left), and against motor noise and the radius within which a drone sees the template (right).}
\label{fig:x5}
\end{figure*}

\begin{figure*}[t]
\centering
\includegraphics[width=\textwidth]{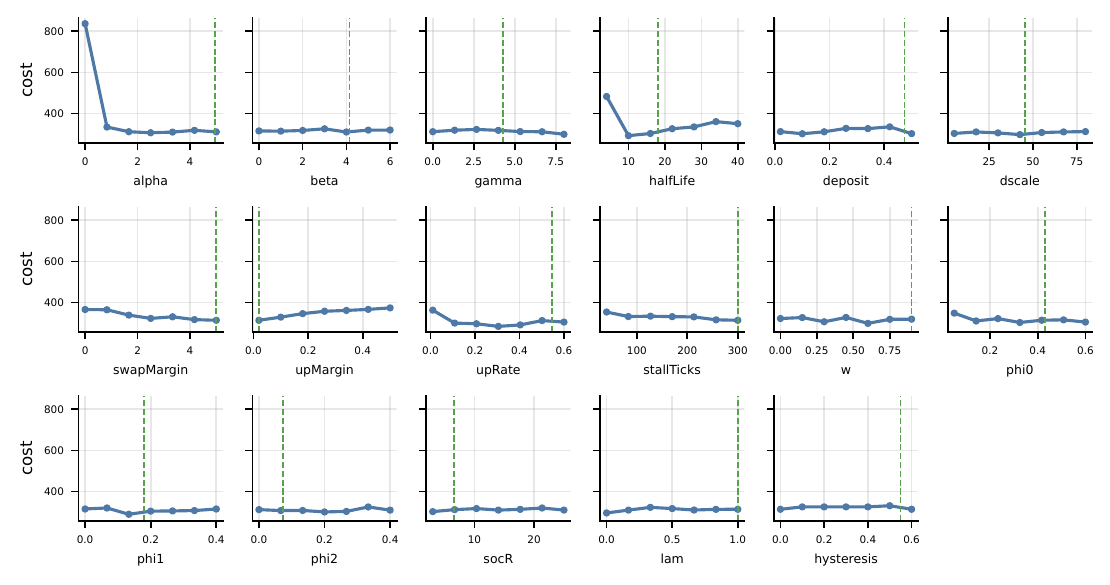}
\caption{One-at-a-time sensitivity around the tuned values (40 runs per point; cost as in tuning, lower is better; dashed line: tuned value).}
\label{fig:x6}
\end{figure*}

%% file: tables/params.tex
\begin{tabular}{lrrrr}\toprule
Parameter & Range & Hand-set & PSO-tuned & Random-search \\\midrule
$\mathtt{alpha}$ & [0, 5] & 2 & 4.94 & 4.87 \\
$\mathtt{beta}$ & [0, 6] & 2 & 4.14 & 3.55 \\
$\mathtt{gamma}$ & [0, 8] & 3 & 4.28 & 7.39 \\
$\mathtt{halfLife}$ & [4, 40] & 12 & 18.2 & 39.9 \\
$\mathtt{deposit}$ & [0.02, 0.5] & 0.15 & 0.474 & 0.328 \\
$\mathtt{dscale}$ & [5, 80] & 20 & 45.5 & 19.2 \\
$\mathtt{swapMargin}$ & [0, 5] & 1 & 5 & 4.91 \\
$\mathtt{upMargin}$ & [0.02, 0.5] & 0.12 & 0.02 & 0.0919 \\
$\mathtt{upRate}$ & [0.01, 0.6] & 0.1 & 0.545 & 0.122 \\
$\mathtt{stallTicks}$ & [40, 300] & 120 & 300 & 205 \\
$\mathtt{w}$ & [0, 0.9] & 0.5 & 0.9 & 0.511 \\
$\mathtt{phi0}$ & [0.05, 0.6] & 0.25 & 0.43 & 0.287 \\
$\mathtt{phi1}$ & [0, 0.4] & 0.1 & 0.179 & 0.267 \\
$\mathtt{phi2}$ & [0, 0.4] & 0.1 & 0.0726 & 0.346 \\
$\mathtt{socR}$ & [3, 25] & 10 & 6.57 & 9.56 \\
$\mathtt{lam}$ & [0, 1] & 0.3 & 1 & 0.496 \\
$\mathtt{hysteresis}$ & [0, 0.6] & 0.25 & 0.549 & 0.425 \\
\bottomrule\end{tabular}

%% file: tables/params_sel.tex
\begin{tabular}{lrrrr}\toprule
 & ACO-PSO & ACO-PSO & CBBA-r & CBBA-r \\
Parameter & reference & best of five & reference & best of five \\\midrule
$\mathtt{alpha}$ & 4.94 & 4.09 & 3.93 & 2.15 \\
$\mathtt{beta}$ & 4.14 & 4.88 & 0.487 & 0 \\
$\mathtt{gamma}$ & 4.28 & 0.43 & 3.92 & 0.864 \\
$\mathtt{halfLife}$ & 18.2 & 19.6 & 19.3 & 16 \\
$\mathtt{deposit}$ & 0.474 & 0.02 & 0.122 & 0.02 \\
$\mathtt{dscale}$ & 45.5 & 65 & 73.5 & 28.8 \\
$\mathtt{swapMargin}$ & 5 & 2.6 & -- & -- \\
$\mathtt{upMargin}$ & 0.02 & 0.02 & 0.0319 & 0.02 \\
$\mathtt{upRate}$ & 0.545 & 0.102 & 0.594 & 0.6 \\
$\mathtt{stallTicks}$ & 300 & 243 & 189 & 300 \\
$\mathtt{w}$ & 0.9 & 0.571 & 0.9 & 0.9 \\
$\mathtt{phi0}$ & 0.43 & 0.191 & 0.43 & 0.43 \\
$\mathtt{phi1}$ & 0.179 & 0.187 & 0.179 & 0.179 \\
$\mathtt{phi2}$ & 0.0726 & 0.304 & 0.0726 & 0.0726 \\
$\mathtt{socR}$ & 6.57 & 3 & 6.57 & 6.57 \\
$\mathtt{lam}$ & 1 & 0.943 & 1 & 1 \\
$\mathtt{hysteresis}$ & 0.549 & 0.555 & 0.183 & 0.591 \\
\midrule meta-seed & 2024 & 2 & 2024 & 1 \\
selected by & PSO & PSO & RS & PSO \\
validation cost & 476.7 & 428.7 & 459.4 & 391.7 \\
\bottomrule\end{tabular}

%% file: tables/x16_configs.tex
\begin{tabular}{llrrrrrrrrr}\toprule
 & & & \multicolumn{2}{c}{easy} & \multicolumn{2}{c}{normal} & \multicolumn{2}{c}{hard} & \multicolumn{2}{c}{brutal} \\
\cmidrule(lr){4-5}\cmidrule(lr){6-7}\cmidrule(lr){8-9}\cmidrule(lr){10-11}
Controller & meta-seed & val.\ cost & succ. & PAR-2 & succ. & PAR-2 & succ. & PAR-2 & succ. & PAR-2 \\\midrule
ACO-PSO & 2024 (ref.) & 476.7 (PSO) & 100 & 104 & 100 & 105 & 100 & 134 & 95 & 554 \\
 & 1 & 429.9 (PSO) & 100 & 119 & 100 & 116 & 100 & 151 & 96 & 403 \\
 & 2$^{b}$ & 428.7 (PSO) & 100 & 133 & 100 & 132 & 100 & 154 & 99 & 367 \\
 & 3 & 442.9 (PSO) & 100 & 127 & 100 & 123 & 100 & 145 & 100 & 287 \\
 & 4 & 439.4 (PSO) & 100 & 137 & 100 & 133 & 100 & 156 & 99 & 332 \\
 & mean of 5 & & 100 & 124 & 100 & 122 & 100 & 148 & 98 & 389 \\\midrule
CBBA w.\ reset & 2024 (ref.) & 459.4 (RS) & 100 & 101 & 100 & 103 & 100 & 136 & 99 & 426 \\
 & 1$^{b}$ & 391.7 (PSO) & 100 & 102 & 100 & 104 & 100 & 147 & 100 & 315 \\
 & 2 & 393.1 (PSO) & 100 & 102 & 100 & 103 & 100 & 143 & 99 & 281 \\
 & 3 & 438.1 (RS) & 100 & 117 & 100 & 123 & 100 & 150 & 100 & 301 \\
 & 4 & 397.4 (PSO) & 100 & 102 & 100 & 104 & 100 & 145 & 100 & 295 \\
 & mean of 5 & & 100 & 105 & 100 & 107 & 100 & 144 & 99 & 324 \\\midrule
\multicolumn{3}{l}{ACO-PSO faster (of 25 pairs), $p$} & \multicolumn{2}{c}{1/25, 0.02} & \multicolumn{2}{c}{3/25, 0.06} & \multicolumn{2}{c}{7/25, 0.31} & \multicolumn{2}{c}{7/25, 0.31} \\
\bottomrule\end{tabular}

%% file: tables/x1.tex
\begin{tabular}{llrrrrr}\toprule
Difficulty & Method & Success\,\% [95\% CI] & Median $T_{95}$ [95\% CI] & PAR-2 & Assign.\ ratio & $p$ vs tuned \\\midrule
easy & Random slot & 100 [97,100] & 109 [108,110] & 109 & 1.05 & $<$0.001 \\
 & Greedy nearest & 100 [97,100] & 116 [113,118] & 116 & 1.03 & $<$0.001 \\
 & CBBA-like & 100 [97,100] & 108 [107,110] & 109 & 1.05 & $<$0.001 \\
 & CBBA with reset & 100 [97,100] & 109 [108,110] & 110 & 1.05 & $<$0.001 \\
 & CBBA with reset (tuned) & 100 [97,100] & 101 [99,104] & 101 & 1.05 & $<$0.001 \\
 & ACO-PSO (hand-set) & 100 [97,100] & 118 [116,120] & 121 & 1.01 & $<$0.001 \\
 & ACO-PSO (tuned) & 100 [97,100] & 103 [102,104] & 104 & 1.01 & -- \\
 & Central oracle (Hungarian) & 100 [97,100] & 94 [89,96] & 93 & 1.00 & $<$0.001 \\
\midrule
normal & Random slot & 100 [97,100] & 113 [111,115] & 113 & 1.06 & $<$0.001 \\
 & Greedy nearest & 100 [97,100] & 118 [117,120] & 119 & 1.02 & $<$0.001 \\
 & CBBA-like & 100 [97,100] & 110 [109,111] & 110 & 1.04 & $<$0.001 \\
 & CBBA with reset & 100 [97,100] & 110 [109,113] & 111 & 1.04 & $<$0.001 \\
 & CBBA with reset (tuned) & 100 [97,100] & 104 [102,105] & 102 & 1.04 & $<$0.001 \\
 & ACO-PSO (hand-set) & 100 [97,100] & 122 [118,126] & 124 & 1.01 & $<$0.001 \\
 & ACO-PSO (tuned) & 100 [97,100] & 104 [103,106] & 105 & 1.01 & -- \\
 & Central oracle (Hungarian) & 100 [97,100] & 95 [91,97] & 95 & 1.00 & $<$0.001 \\
\midrule
hard & Random slot & 100 [97,100] & 168 [155,193] & 250 & 2.73 & $<$0.001 \\
 & Greedy nearest & 100 [97,100] & 280 [242,307] & 307 & 3.03 & $<$0.001 \\
 & CBBA-like & 84 [77,90] & 139 [137,141] & 600 & 1.34 & $<$0.001 \\
 & CBBA with reset & 100 [97,100] & 136 [127,151] & 150 & 1.63 & $<$0.001 \\
 & CBBA with reset (tuned) & 100 [97,100] & 128 [119,145] & 135 & 1.41 & 0.323 \\
 & ACO-PSO (hand-set) & 100 [97,100] & 148 [136,160] & 161 & 1.11 & $<$0.001 \\
 & ACO-PSO (tuned) & 100 [97,100] & 140 [120,144] & 134 & 1.08 & -- \\
 & Central oracle (Hungarian) & 100 [97,100] & 96 [92,98] & 96 & 1.00 & $<$0.001 \\
\midrule
brutal & Random slot & 62 [53,70] & 684 [548,945] & 1629 & 2.30 & $<$0.001 \\
 & Greedy nearest & 89 [82,94] & 583 [448,679] & 893 & 2.42 & $<$0.001 \\
 & CBBA-like & 5 [2,10] & 204 [188,221] & 2860 & 1.17 & $<$0.001 \\
 & CBBA with reset & 89 [82,94] & 422 [205,539] & 758 & 1.64 & $<$0.001 \\
 & CBBA with reset (tuned) & 99 [95,100] & 183 [169,396] & 427 & 1.49 & 0.680 \\
 & ACO-PSO (hand-set) & 72 [63,79] & 631 [198,884] & 1307 & 1.44 & $<$0.001 \\
 & ACO-PSO (tuned) & 98 [93,99] & 190 [165,272] & 489 & 1.23 & -- \\
 & Central oracle (Hungarian) & 100 [97,100] & 100 [100,101] & 100 & 1.00 & $<$0.001 \\
\midrule
\bottomrule\end{tabular}

%% file: tables/x3.tex
\begin{tabular}{lrrrrrrr}\toprule
Event & \shortstack{Central oracle\\(Hungarian)} & \shortstack{ACO-PSO\\(tuned)} & \shortstack{ACO-PSO\\(hand-set)} & \shortstack{CBBA with reset} & \shortstack{CBBA with reset\\(tuned)} & \shortstack{CBBA-like} & \shortstack{Greedy nearest} \\\midrule
lose 10\% & 4 / 100 / 1.00 & 1 / 100 / 0.98 & 1 / 100 / 0.96 & 1 / 100 / 0.96 & 1 / 100 / 0.99 & 1 / 100 / 0.91 & 1 / 100 / 0.91 \\
lose 25\% & 18 / 100 / 1.00 & 178 / 95 / 0.95 & 121 / 99 / 0.96 & 94 / 98 / 0.96 & 116 / 100 / 0.98 & -- / 0 / 0.77 & -- / 0 / 0.76 \\
lose 50\% & 27 / 100 / 1.00 & 186 / 96 / 0.96 & 146 / 85 / 0.95 & 114 / 88 / 0.95 & 128 / 96 / 0.98 & -- / 0 / 0.50 & -- / 0 / 0.49 \\
lose 75\% & 30 / 100 / 1.00 & 197 / 71 / 0.91 & 164 / 38 / 0.90 & 124 / 30 / 0.88 & 150 / 88 / 0.92 & -- / 0 / 0.29 & -- / 0 / 0.26 \\
lose upper half & 48 / 100 / 1.00 & 202 / 99 / 0.97 & 210 / 90 / 0.94 & 143 / 92 / 0.94 & 142 / 100 / 0.97 & -- / 0 / 0.38 & -- / 0 / 0.38 \\
4 cascading losses of 15\% & 19 / 100 / 1.00 & 157 / 96 / 0.97 & 104 / 74 / 0.94 & 84 / 68 / 0.94 & 110 / 100 / 0.98 & -- / 0 / 0.56 & -- / 0 / 0.55 \\
12 drones join & 68 / 100 / 1.00 & 81 / 100 / 1.00 & 89 / 100 / 1.00 & 92 / 100 / 1.00 & 82 / 100 / 1.00 & 92 / 100 / 1.00 & 102 / 100 / 1.00 \\
\midrule
\multicolumn{8}{l}{\textit{Continuous churn (death rate 0.2\%/drone/tick, balanced joins): mean picture quality over ticks 200--700}}\\
priority fill & 86\% & 84\% & 84\% & 85\% & 84\% & 48\% & 82\% \\
excess Chamfer (units) & 2.46 & 1.74 & 1.68 & 1.76 & 1.77 & 5.87 & 1.75 \\
\bottomrule\end{tabular}

%% file: tables/x1_cost.tex
\begin{tabular}{lrrrr}\toprule
Method & Msgs/drone/tick & Table entries/drone & Path length & Conflicts \\\midrule
Random slot & 10.7 & 34.0 & 9695 & 44 \\
Greedy nearest & 10.7 & 34.0 & 9626 & 339 \\
CBBA-like & 10.7 & 34.0 & 9665 & 368 \\
CBBA with reset & 10.7 & 34.0 & 9663 & 375 \\
CBBA with reset (tuned) & 10.7 & 34.0 & 9690 & 489 \\
ACO-PSO (hand-set) & 10.7 & 34.0 & 5801 & 713 \\
ACO-PSO (tuned) & 10.6 & 34.0 & 9626 & 124 \\
Central oracle (Hungarian) & 0.0 & 0.0 & 9598 & 0 \\
\bottomrule\end{tabular}

%% file: tables/x7.tex
\begin{tabular}{llrrrrrr}\toprule
 & & \multicolumn{3}{c}{Safety layer ON} & \multicolumn{3}{c}{Safety layer OFF} \\
Method & noise $\times$ & runs & overlap pair-ticks & min sep & runs & overlap pair-ticks & min sep \\\midrule
ACO-PSO (tuned) & 1 & 60 & 0 & 2.10 & 60 & 38324 & 0.02 \\
 & 2 & 60 & 0 & 2.16 & 60 & 31221 & 0.01 \\
 & 4 & 60 & 0 & 2.20 & 60 & 23891 & 0.00 \\
Central oracle (Hungarian) & 1 & 60 & 0 & 2.08 & 60 & 20065 & 0.05 \\
 & 2 & 60 & 0 & 2.17 & 60 & 17817 & 0.04 \\
 & 4 & 60 & 0 & 2.35 & 60 & 12893 & 0.03 \\
CBBA-like & 1 & 60 & 0 & 2.07 & 60 & 31923 & 0.01 \\
 & 2 & 60 & 0 & 2.09 & 60 & 55845 & 0.01 \\
 & 4 & 60 & 0 & 2.12 & 60 & 131415 & 0.00 \\
Random slot & 1 & 60 & 0 & 2.07 & 60 & 36602 & 0.01 \\
 & 2 & 60 & 0 & 2.17 & 60 & 34425 & 0.01 \\
 & 4 & 60 & 0 & 2.16 & 60 & 28783 & 0.01 \\
\bottomrule\end{tabular}

%% file: tables/x11.tex
\begin{tabular}{lrrrr}\toprule
 & \multicolumn{2}{c}{lose 25\%} & \multicolumn{2}{c}{lose 50\%} \\
\cmidrule(lr){2-3}\cmidrule(lr){4-5}
Controller & ticks to PF$\ge$0.9 & PF$_{400}$ & ticks to PF$\ge$0.9 & PF$_{400}$ \\\midrule
Central oracle (Hungarian) & 64 (100\%) & 1.00 & 66 (100\%) & 1.00 \\
ACO-PSO (tuned) & 37 (100\%) & 0.99 & 174 (98\%) & 0.93 \\
CBBA with reset & 48 (100\%) & 0.98 & 100 (96\%) & 0.95 \\
CBBA with reset (tuned) & 62 (100\%) & 0.98 & 114 (99\%) & 0.96 \\
CBBA-like & 136 (1\%) & 0.79 & -- & 0.57 \\
\bottomrule\end{tabular}

%% file: tables/x12.tex
\begin{tabular}{lrrr}\toprule
Motion law (150 drones) & Success \% & Median $T_{95}$ [95\% CI] & $p$ vs full \\\midrule
full controller & 100 & 244 [219,279] & -- \\
proportional motion (no PSO) & 100 & 252 [218,298] & 0.073 \\
without personal/social terms & 100 & 236 [213,272] & 0.249 \\
without crowding term & 100 & 238 [221,270] & 0.808 \\
\bottomrule\end{tabular}

%% file: tables/x4.tex
\begin{tabular}{lrrr}\toprule
Method & exponent $b$ ($T\propto N^{b}$) & 95\% CI & success at $N=300$ \\\midrule
Central oracle (Hungarian) & 0.63 & [0.56, 0.69] & 100\% \\
ACO-PSO (tuned) & 0.63 & [0.57, 0.68] & 100\% \\
CBBA with reset (tuned) & 0.58 & [0.54, 0.61] & 67\% \\
CBBA-like & 0.59 & [0.55, 0.63] & 54\% \\
\bottomrule\end{tabular}

%% file: tables/x9_shapes.tex
\begin{tabular}{lrrrrrrrr}\toprule
 & disc & ring & arrow & cross & twoBlobs & letterA & spiral & wave \\\midrule
CBBA-like (normal) & 100 & 100 & 100 & 100 & 100 & 100 & 100 & 100 \\
 (hard) & 100 & 100 & 100 & 100 & 95 & 100 & 50 & 100 \\
 (brutal) & 0 & 5 & 10 & 0 & 0 & 0 & 0 & 0 \\
\midrule
CBBA with reset (normal) & 100 & 100 & 100 & 100 & 100 & 100 & 100 & 100 \\
 (hard) & 100 & 100 & 100 & 100 & 100 & 100 & 100 & 100 \\
 (brutal) & 0 & 100 & 100 & 30 & 100 & 100 & 75 & 100 \\
\midrule
CBBA with reset (tuned) (normal) & 100 & 100 & 100 & 100 & 100 & 100 & 100 & 100 \\
 (hard) & 100 & 100 & 100 & 100 & 100 & 100 & 100 & 100 \\
 (brutal) & 10 & 100 & 100 & 95 & 100 & 100 & 100 & 100 \\
\midrule
ACO-PSO (tuned) (normal) & 100 & 100 & 100 & 100 & 100 & 100 & 100 & 100 \\
 (hard) & 100 & 100 & 100 & 100 & 100 & 100 & 100 & 100 \\
 (brutal) & 80 & 100 & 100 & 100 & 100 & 100 & 100 & 100 \\
\midrule
Central oracle (Hungarian) (normal) & 100 & 100 & 100 & 100 & 100 & 100 & 100 & 100 \\
 (hard) & 100 & 100 & 100 & 100 & 100 & 100 & 100 & 100 \\
 (brutal) & 100 & 100 & 100 & 100 & 100 & 100 & 100 & 100 \\
\midrule
\bottomrule\end{tabular}